\documentclass[runningheads]{llncs}

\usepackage{eccv}

\newcommand\samethanks[1][\value{footnote}]{\footnotemark[#1]}

\usepackage{eccvabbrv}

\usepackage{graphicx}
\usepackage{booktabs}
\usepackage{siunitx}
\usepackage{float}

\newcommand{\R}{\mathbb{R}}
\let \bs=\mathbf
\let \set=\mathcal

\def \mean {\textup{mean}}
\def \sync {\textup{sync}}

\def \data {\textup{data}}
\def \reg {\textup{reg}}

\def \reg {\textup{reg}}

\def \path {\mathit{path}}

\let \set = \mathcal
\let \bs = \boldsymbol

\usepackage[accsupp]{axessibility}  

\usepackage{hyperref}

\usepackage{orcidlink}

\begin{document}

\title{GenSP: Consistent Spherical Parameterization via Learning Shape Generative Models} 

\titlerunning{GenSP: Consistent Spherical Parameterizations}

\author{Sai Karthikey Pentapati\inst{1}\thanks{Equal contribution.} \and
Shashank Gupta\inst{1}\samethanks \and
Rajesh Sureddi\inst{1} \and 
Yuezhi Yang\inst{1} \and
Alan C. Bovik\inst{2} \and
Qixing Huang\inst{1}}

\authorrunning{S.K.~Pentapati et al.}

\institute{The University of Texas at Austin \\
\email{\{ps.karthik,shashank.gupta\}@utexas.edu}\\ \and
University of Colorado Boulder\\
}
\maketitle

\begin{abstract}

We introduce GenSP, a data-driven framework that learns consistent spherical parameterizations across a collection of genus-0 shapes. 
Instead of optimizing the parameterization of each shape independently, our method learns a neural generative model that predicts a continuous mapping from the unit sphere to shapes in a dataset. 
Under this formulation, spherical parameterizations are obtained through the inverse mappings of the learned generator, which encourages similar shapes to share consistent parameterizations. To make this formulation practical, we address several key challenges in learning such a generative model. First, we introduce a continuous neural deformation model that predicts surface points from sphere coordinates and latent shape codes, avoiding discretization artifacts common in mesh-based formulations. Second, we augment the training space with intermediate shapes that bridge the sphere and input shapes, allowing the model to learn meaningful deformations across a heterogeneous shape collection. Third, we compute reliable initial correspondences by propagating mappings along a spanning tree of training shapes in the latent space. Experiments on the ShapeNet \cite{chang2015shapenet} dataset demonstrate that our approach significantly reduces geometric distortion and improves cross-shape consistency compared with state-of-the-art spherical parameterization methods.
\end{abstract}
\section{Introduction}

\begin{figure}[t]
\centering
\includegraphics[width=\textwidth]{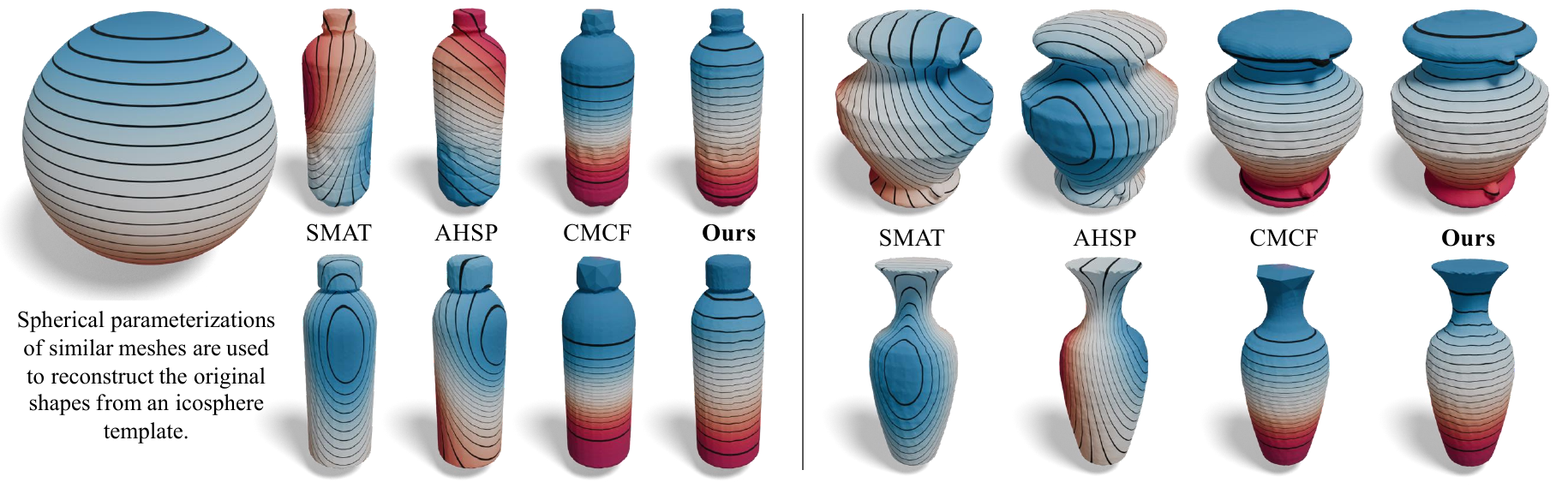}
\caption{\textbf{Comparison of spherical parameterizations with existing approaches.}  A texture defined on an icosphere is mapped onto each shape through the spherical parameterization to visualize the consistency of the mappings across shapes. Existing approaches either produce inconsistent parameterizations across shapes (SMAT~\cite{DBLP:journals/cgf/SchmidtPK23}, AHSP~\cite{Hu-2017-AHSP}) or exhibit high geometric distortion in thin or high-curvature regions (CMCF~\cite{DBLP:journals/cgf/KazhdanSB12}). In contrast, GenSP produces consistent, feature-preserving  and accurate parameterizations across the shape collection.}
\label{Figure:Teaser}
\vspace{-0.2in}
\end{figure}

Computing spherical parameterizations of genus-0 surfaces is a fundamental problem in geometry processing and visual computing~\cite{DBLP:journals/ftcgv/ShefferPR06,DBLP:conf/siggraph/HormannLS07,DBLP:journals/cgf/WilliamsonM25, hns}. 
Given a genus-0 surface, spherical parameterization computes a bijective mapping between the surface and the unit sphere, providing a canonical domain for applications such as texture transfer, shape correspondence, and geometric analysis. Compared with planar parameterization methods that map surfaces to the UV plane, spherical parameterization does not require cutting the surface. Such cuts introduce discrete optimization decisions that are difficult to solve and often lead to additional distortion in the parameterization. Most existing approaches~\cite{10.1145/1201775.882276,10.1145/882262.882274} compute spherical parameterizations for each surface independently.  However, a fundamental challenge is that there exist infinitely many spherical parameterizations for a genus-0 surface. 
Even when the mapping is restricted to be conformal, the space of valid parameterizations forms a M{\"{o}}bius group~\cite{10.1145/1531326.1531378}. 
As a result, two geometrically similar shapes may obtain very different parameterizations when using existing methods, as shown in Figure~\ref{Figure:Teaser}.

In this paper, we introduce a data-driven approach for computing consistent spherical parameterizations across a collection of genus-0 shapes. 
Our approach learns a generative model that continuously deforms the unit sphere to each shape in the dataset. Under this formulation, spherical parameterizations arise naturally as the inverse mappings of the learned generator. To ensure consistent parameterizations across shapes, we formulate common objectives of spherical parameterization as regularization losses on the generative model. 
These losses encourage the generator to produce smooth deformations, which lead to similar spherical parameterizations among neighboring shapes in the dataset. 
We further introduce a deformation regularization term that effectively minimizes geometric distortion in the resulting parameterizations.

Although the conceptual idea of this approach is straightforward, there are three fundamental challenges. The first challenge is the representation of the generative model. The straightforward approach of learning a mesh generative model is not ideal due to large variations in conformal factors of the underlying parameterization, leading to unevenly large and tiny triangles during parameterization. To address this issue, we introduce a continuous neural deformation model that maps points on a sphere to corresponding surface points on the target shape.

The second challenge is that, to date, most shape generative models with fixed topology have only been applied to relatively limited shape spaces, e.g., humans/animals~\cite{SMPL:2015,Zuffi:CVPR:2017}. In many cases, the underlying sub-space does not contain the sphere, making it difficult to produce smooth latent space trajectories from a sphere to the shapes in a dataset. 
To address this issue, we propose augmenting the shape space with additional shapes that bridge the sphere and the input shapes. This is achieved by running existing flow-based models~\cite{DBLP:journals/cgf/KazhdanSB12} to deform a shape into a sphere and adding suitable intermediate shapes as training shapes to learn the generative model. 
The training dataset, augmented with intermediate shape samples, samples the distribution of a larger genus-0 shape space that contains both the sphere and the input shapes in a more natural manner.  

The final challenge is to compute the correspondences between the sphere and each input shape to initialize our neural deformation model, since learning this generator from scratch is prone to getting trapped in local minima. We find that doing so directly is quite difficult, particularly when the deformation between the sphere and an input shape is large. We address this by finding a path of training (input and augmented) shapes that gradually deforms the sphere into each input shape. 
To this end, we learn a point cloud encoder and an implicit decoder using an auto-encoder loss with additional regularization losses. The first loss controls the distribution of latent codes, wherein the latent code of the sphere is forced to lie at the origin of the latent space, and the latent codes of two shapes are close if and only if they are similar. The second loss regularizes the shape space defined by the shape decoder, so that intermediate shapes between two adjacent input shapes are deformation preserving, prioritizing high-quality inter-shape correspondences. 
We then construct a spanning tree among the latent codes of the training shapes, with the latent code of the sphere at the root.
Then, obtaining correspondences between a complex training shape and the sphere is done by composing correspondences between shapes associated with each edge along the shortest path between the given shape and the sphere.
These point-to-point correspondences are consistent between similar training shapes and are used to initialize the neural-deformation based mesh generative model.
Finally, we regularize the model to minimize the isometric distortion in the generated mappings.


We tested our approach on the Genus-0 subset of the ShapeNet \cite{chang2015shapenet} dataset. The experimental results show that our approach  yields parameterizations with significantly improved inter-shape consistency compared to state-of-the-art spherical parameterizations while simultaneously having low geometric distortions. 
\section{Related Work}

\subsection{Spherical Mesh Parameterization}

Existing spherical parameterizations have evolved through four generations. The first generation reduces the problem to the planar case. \cite{Haker:2000:CSP} first cut out one triangle to serve as a boundary, then parameterize the resulting open mesh over the unit triangle using any planar parameterization method, and finally use the inverse stereographic projection to map the plane to the sphere. \cite{DBLP:journals/vc/ChenXC25} is a modern variant of this approach. The main problem with these methods is the severe distortion of large regions. The key idea of the second generation~\cite{Shapiro1998Polygon,DasGoodrich1997,10.1145/882262.882274, DBLP:journals/cgf/SchmidtPK23, Hu-2017-AHSP} is to perform mesh simplification into a tetrahedron, map the tetrahedron to the sphere, and revert the procedure to compute the spherical parameterization of each intermediate mesh in the simplification process. Although this method is efficient, it is difficult to optimize the parameterization due to its greedy nature. Additionally, the bottom-to-top approach makes it challenging to preserve the consistency between parameterizations of two similar shapes.
The third generation consists of various global optimization methods. The seminal work of~\cite{10.1145/1201775.882276} connects discrete spherical parameterizations to spectral graph theory and computes spherical parameterization via a generalized Laplacian. However, this formulation is non-convex and has many local minima. Several other papers~\cite{Alexa00,Kobbelt1999ASW, 10.1016/j.gmod.2014.03.016} solve various global optimization relaxations, but all have issues of local minima. The fourth generation models spherical parameterizations as geometric flows~\cite{DBLP:journals/cgf/KazhdanSB12,10.1145/2461912.2461986,DBLP:journals/cgf/BadenCK18}, which gradually deform a given shape into the sphere. These approaches enjoy theoretical guarantees and improved consistency between parameterizations. However, these methods often yield extremely poor isometry factors (see Figure \ref{Figure:Teaser}). Our approach can be considered a variant of the flow-based approaches, in which the flow from each shape to the sphere is not determined by a PDE but by data, i.e., a path of intermediate shapes. This approach greatly enhances the stability of spherical parameterization. 
\vspace{-0.1in}
\subsection{Shape Correspondences}

Spherical parameterization can be formulated as a shape matching problem, which computes dense correspondences between vertices of a triangular mesh and the sphere. However, existing shape-matching approaches do not apply to the problem of matching a sphere with an arbitrary genus-0 shape. A breakthrough result in this domain is the functional map representation~\cite{10.1145/2185520.2185526,10.1145/3588432.3591518}. Due to their simple algebraic representation, functional maps have been widely adopted in the deep learning era~\cite{DBLP:conf/cvpr/ZhuravlevLG25,Pierson_2025_ICCV,DBLP:conf/cvpr/DonatiSO20,DBLP:conf/iccv/LitanyRRBB17,DBLP:conf/3dim/AttaikiPO21,DBLP:conf/cvpr/DonatiCO22}. Several other approaches~\cite{DBLP:conf/eccv/GroueixFKRA18,DBLP:conf/cvpr/EisenbergerNKLN21,DBLP:journals/cgf/EisenbergerLC19,DBLP:conf/nips/TrappoliniCMMMR21} also introduced deep neural networks for shape matching. Compared to non-deep-learning-based approaches,  learning-based methods relax the constraint that the input shapes should be similar to each other. However, they are still not applicable to matching a sphere shape with an arbitrary genus-0 shape due to a lack of identifiable keypoints on the sphere.

One promising approach to extend the range of input shapes is to perform joint shape matching~\cite{DBLP:journals/cgf/NguyenBWYG11,Kim:2012:ECM,DBLP:journals/tog/HuangWG14}, where the map between two shapes is the composition of pairwise maps along a path of similar shape pairs. Joint shape matching typically enforces the cycle-consistency constraint, which is effective for pruning incorrect maps computed between pairs of shapes. Several papers have studied how to enforce this constraint using convex~\cite{DBLP:journals/cgf/HuangG13}, non-convex~\cite{zhou2015multi}, and spectral techniques~\cite{DBLP:conf/nips/PachauriKS13,DBLP:conf/nips/ShenHSS16}. The most relevant to our approach is GenCorres~\cite{DBLP:conf/iclr/0005HSBH24}, which learns mesh generative models to establish consistent correspondences. 
However, learning a mesh generative model requires high-quality initial correspondences, which are difficult to obtain. Moreover, a mesh generative model is not applicable in our setting due to large variations of conformal factors between the underlying shapes. This paper addresses these two issues by studying the problems of learning a shape generative model from a heterogeneous shape collection of diverse genus-0 shapes and encoding the shape generative model as neural fields. 

\vspace{-0.1in}
\subsection{Shape Generative Models}
GenSP is based on recent advances in generative models of deformable shapes. Early work on this topic employs explicit shape representations, such as distance matrices~\cite{DBLP:conf/eccv/CosmoNHKR20} between all pairs of points and triangular meshes~\cite{DBLP:conf/eccv/ZhouBP20,DBLP:conf/eccv/RanjanBSB18,DBLP:conf/iccv/BouritsasBPZB19,DBLP:conf/nips/ZhouWLCYSLS20}. These approaches require consistently meshed training shapes as input for learning the generative model. The idea of alternating between computing consistent correspondences and learning an explicit generative model requires good initializations, which this paper addresses. 

Several other papers studied learning an implicit shape generative model that bypasses the requirement of correspondences among the input data. Motivated by DeepSDF~\cite{DBLP:conf/cvpr/ParkFSNL19}, SALD~\cite{DBLP:conf/iclr/AtzmonL21} learns an MLP to encode an implicit shape generative model for deformable shapes, e.g., humans and animals. This approach does not apply to training shapes with large pose variations. Several papers introduced regularization losses for implicit generative models. Among them, the killing vector field approach~\cite{DBLP:conf/cvpr/SlavchevaBCI17}, which provides weak regularizations, has been adopted in several papers~\cite{DBLP:conf/cvpr/ChibaneAP20,DBLP:journals/corr/abs-2108-08931,DBLP:conf/cvpr/PumarolaCPM21,DBLP:conf/iccv/TretschkTGZLT21}. GenCorres~\cite{DBLP:conf/iclr/0005HSBH24} and GenAnalysis~\cite{DBLP:journals/tog/YangYNHGH25} introduce a stronger regularization, which is based on discretizing an implicit surface generator into a mesh generator locally by computing dense correspondences of mesh vertices on adjacent implicit surfaces. This allows us to employ prior work that enforces deformation losses on meshes~\cite{DBLP:conf/iccv/HuangHSZJB21,10.1145/3618371} to regularize the implicit generator. This approach is flexible because it allows us to employ a reduced deformation model that enables stronger and more efficient regularizations, which we will explore in this paper. 

\vspace{-0.1in}
\subsection{Neural Deformation Models}

Our approach uses classical shape deformation models such as as-rigid-as-possible~\cite{sorkine2007arap} and as-conformal-as-possible~\cite{yoshiyasu2014conformal}. Unlike the standard formulations that do not admit a closed-form expression, we employ the linear approximation~\cite{hwag-sduma-09}, which applies to shapes that are close to each other and admits a closed-form expression. 

The application of this deformation model to regularize generative models has been studied in the literature, e.g., ARAPReg~\cite{DBLP:conf/iccv/HuangHSZJB21}, GeoLatent~\cite{10.1145/3618371} and GenCorres~\cite{DBLP:conf/iclr/0005HSBH24}. These methods introduce regularization losses to minimize the deformation between adjacent synthetic shapes. Although our approach also follows this paradigm, the key difference is that we normalize the deformation using the absolute shape displacements, addressing the issue of irregular distributions of shape latent codes. 

Several other papers have studied neural deformation models, e.g., NJField~\cite{10.1145/3528223.3530141}, LLDPD~\cite{10.1145/3687968}, and TutteNet~\cite{Sun_2024_CVPR}. A common scheme among these models is to employ neural networks to predict the deformation field, e.g., displacements of deformation nodes and deformation gradients. The focus is on shapes that undergo large deformations. However, neural shape deformation models have generalization issues. In contrast, our deformation model is between adjacent shapes, and deformations between different shapes are computed using a generative model that implicitly composes deformations between adjacent shapes.

\section{Problem Statement and Approach Overview}


\subsection{Problem Statement}
\label{Subsec:PS}

The input consists of a collection of triangular meshes $\set{S} = \{S_0,\cdots S_N\}$. Each mesh is closed and has genus-$0$. Here $S_0$ is a mesh of the unit sphere with $n$ vertices. $S_1,\cdots, S_N$ are actual training meshes. As a starting point, we assume that these meshes are roughly normalized in some shared ambient shape. However, we do not assume that they are consistently meshed. 

Our goal is to find, for each shape $S_i$, a mapping $m_i: S_i\rightarrow S_0$ to the unit sphere $S_0$. $m_i$ has the following properties: 1) $m_i$ is bijective, i.e., the image $m_i(S_i)$ covers $S_0$ and does not have self-intersections. 2) $m_i$ has minimal isometric deformations. 3) For similar shapes $S_i$ and $S_j$, the mappings $m_i$ and $m_j$ are also similar.

\vspace{-0.1in}
\subsection{Approach Overview}
\label{Subsec:AO}

Our approach learns a neural shape generative model of the inverse mapping from the unit sphere to each shape in the genus-$0$ shape space. 
Denote $\set{Z} \cong \R^d$ as the latent space of genus-0 shapes. 
The generative model is denoted as $\bs{g}^{\theta}(\bs{z},\cdot): \R^3\rightarrow \R^3$, where $\bs{g}^{\theta}(\bs{z},\bs{p})$ is the corresponding surface point of a point on a sphere $\bs{p}\in \mathbb{S}^2$. 
We consider four objectives: 
\begin{enumerate}
    \item The latent code of $S_0$ is at the origin, i.e., $\bs{g}^{\theta}(\bs{0},\bs{p}) \in S_0, \forall \bs{p}\in \mathbb{S}^2$.
    \item $\bs{g}^{\theta}(\bs{z}_i,\bs{p})\in S_i, \forall \bs{p}\in \mathbb{S}^2$, where $\bs{z}_i$ is the latent code of shape $S_i$.
    \item The Jacobian $\frac{\partial \bs{g}^{\theta}(\bs{z},\bs{p})}{\partial \bs{z}}, \forall \bs{p}\in \mathbb{S}^2$ follows a regularized as-conformal-as possible deformation prior.
    \item The geometric distance between two shapes $S_i$ and $S_j$ correlates with the Euclidean distance between $\bs{z}_i$ and $\bs{z}_j$. 
\end{enumerate}

There are two fundamental challenges to learning the generative model described above. First, the sphere $S_0$ is usually not in the underlying shape space of $\set{S}=\{S_1,\cdots, S_N\}$, making it very difficult to learn a shape generative model that fits $\set{S}$ and provides meaningful interpolations between $S_0$ and each $S_i$. To address this challenge, our key idea is to augment the input shapes with additional shapes $S_{N+1},\cdots, S_{\overline{N}}$ by deforming each input shape $S_i$ into the sphere $S_0$ using an off-the-shelf method to obtain the augmented dataset $\overline{\set{S}}$. Second, we need consistent correspondences between the sphere and the training shapes to define a meaningful data term to learn the shape generative model. Without correspondences, minimizing the Chamfer distance or Earth-Mover distance between the synthetic mesh and each training shape can easily result in getting trapped in local minima.  To address this challenge, we propose to introduce an implicit generative model $f^{\phi}:\R^3\times \R^d \rightarrow \R$ to facilitate the computation of initial correspondences for learning this shape generative model.

\begin{figure*}[t!]
\includegraphics[width=\textwidth]{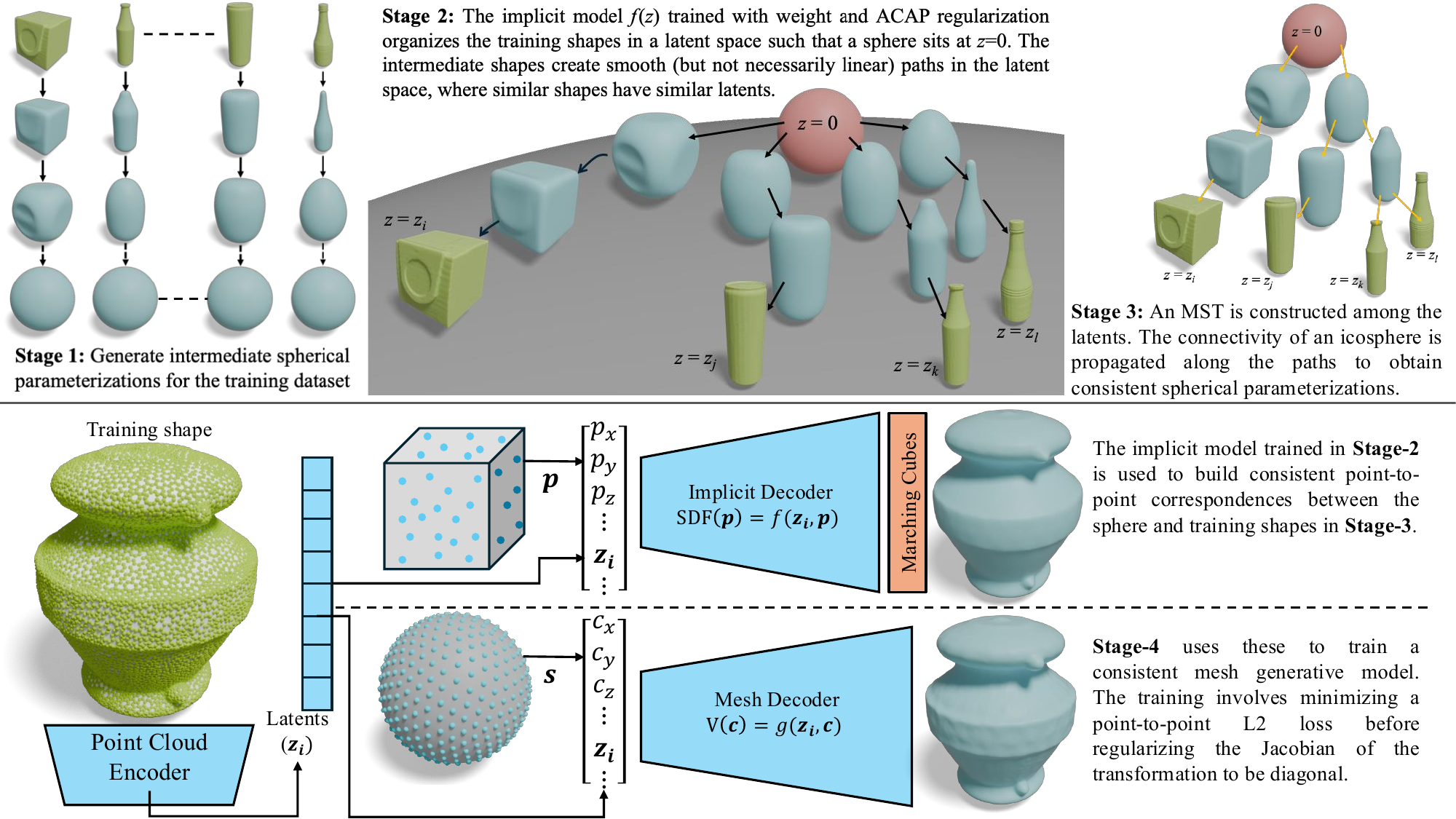}
\caption{An overview of the four stages of GenSP.}
\label{Figure:Overview} 
\vspace{-0.1in}   
\end{figure*}

As shown in Figure~\ref{Figure:Overview}, GenSP proceeds in four stages. The first stage applies an off-the-shelf flow method~\cite{DBLP:journals/cgf/KazhdanSB12} to deform each shape $S_i$ into the sphere. These intermediate shapes are used to augment the training shapes. 

The second stage learns an implicit shape generator $f^{\phi}:\R^d\times \R^3\rightarrow \R$ together with a point cloud encoder $e^{\gamma}:\overline{\set{S}}\rightarrow \set{Z}$ that maps a shape into the latent space using an auto-encoder loss. We add an additional regularization loss on $e^{\gamma}$ to enforce that similar shapes have close-by latent codes. The decoder $f^{\phi}$ will be used to generate intermediate shapes between pairs of adjacent training shapes to propagate inter-shape correspondences. Therefore, we introduce a deformation regularization loss in $f^{\phi}$ to improve the quality of these intermediate shapes. 

Based on the learned implicit shape generator, the third stage constructs a spanning tree among the training shapes where $S_0$ is the root. Starting from the root, we apply template-based shape registration using intermediate shapes generated by the implicit generator $f^{\phi}$ to compute dense correspondences between $S_0$ and each training shape. The resulting correspondences are used to initialize the shape generator $\bs{g}^{\theta}$, which is trained using an auto-encoder loss. 

The last stage fine-tunes the shape generator $\bs{g}^{\theta}$ and the encoder together $e^{\gamma}$ together, where we introduce a deformation regularization term on the shape generator $\bs{g}^{\theta}$ to optimize the final spherical mesh parameterizations. 

\section{Approach}


\subsection{Stage I: Data Augmentation}

The first stage augments the training data $S_0,\cdots, S_N$ with suitable genus-0 shapes $S_{N+1},\cdots, S_{\overline{N}}$ that bridge $S_0$ and $S_1,\cdots, S_N$, as shown in Figure~\ref{Figure:Overview}. We achieve this goal by using an off-the-shelf flow-based method to deform each input shape $S_i$ into the sphere shape $S_0$ and sample intermediate shapes. We chose CMCF~\cite{DBLP:journals/cgf/KazhdanSB12}, which has a stable open-source implementation. CMCF has the step-size as a critical hyper-parameter, and no fixed value is optimal among all input shapes. Therefore, we need to define a scoring function for intermediate shapes to decide which intermediate shapes to use for augmentation.

For a given hyper-parameter $m$ and one shape $S_i$, denote by $S_i^m(t), 0\leq t \leq 1$ the result of $m$ such that $S_i^m(0) = S_i$, and $S_i^m(1)$ converges to the sphere or becomes degenerate. We define the scoring function as
\begin{equation}
s(m,S_i) = d^2(S_0, S_i^m(1)) + \eta \int_{0}^{1}e_{\textup{deform}}(S_i^m(t), S_i) dt    
\label{Eq:Data:Loss:Stage:I}
\end{equation}
where the first term measures the squared Hausdorff distance between $S_0$ and $S_i^m(1)$, and the second term integrates the deformation between $S_i^m(t)$ and $S_i$. For two meshes $\bs{p}\in \R^{3n}$ and $\bs{q}\in \R^{3n}$ each with $n$ vertices that share the same topology, their deformation is defined as 
\begin{equation}
e_{\textup{deform}}(\bs{p}, \bs{q}) = \sum\limits_{i=1}^{n}\min\limits_{R_i\in \textup{SO}(3)}\sum\limits_{j\in \set{N}_i}\|R_i(\bs{p}_i-\bs{p}_j)-(\bs{q}_i-\bs{q}_j)\|^2 
\end{equation}
where $\set{N}_i$ is the 1-ring neighborhood of the vertex $i$.

For each $S_i$, we pick the hyper-parameter $m_i^{\star}$ that has the smallest $s(m_i^{\star},S_i)$. When $s(m_i^{\star},S_i) < 2\delta$, where $\delta$ is the median of $s(m_i^{\star},S_i)$ among all input shapes, we add $S_i^{m_i^{\star}}(t)$ to the training shapes, where $t$ is randomly sampled in $(0,1)$. The total number of training shapes $\overline{N} = 24N$, i.e., we place 23 samples in $t$.

\subsection{Stage II: Autoencoder for Learning the Shape Space}

The second stage learns a point cloud encoder $e^{\gamma}$ and an implicit generator $f^{\phi}$ by combining an auto-encoder data loss and two regularization losses on $e^{\gamma}$ and $f^{\phi}$, respectively. The data loss fits $f^{\phi}$ to the training shapes:
\begin{equation}
l_{\textup{data}}(\phi, \gamma) := \sum\limits_{i=0}^{\overline{N}} \sum\limits_{(\bs{p},d)\in \set{D}(S_i)} \big(f^{\phi}(\bs{p},e^{\gamma}(S_i))-d\big)^2
\label{Eq:Data:Loss:Stage:II}    
\end{equation}
where $\set{D}(S_i)$ collects signed-distance function (SDF) samples (cf.~\cite{DBLP:conf/iclr/AtzmonL21}) on-and-off the surface of $S_i$ to learn the implicit decoder.

The first regularization loss is enforced on $e^{\gamma}$ to ensure that similar training shapes remain close to each other in the latent space. This is important when using latent codes to construct a spanning tree to initialize the correspondences between $S_0$ and each training shape $S_i$. Here, we directly regularize the magnitude of $\gamma$:
\begin{equation}
l_{\textup{reg,I}} (\gamma) = \rho(\|\gamma\|)
\label{Eq:Regu:Loss:I:Stage:I}
\end{equation}
where $\rho$ is chosen as the $L^1$ norm.

The second regularization loss is enforced on the implicit shape generator so that the intermediate shapes that interpolate adjacent synthetic shapes are deformation-preserving. As we will linearly interpolate latent codes of adjacent synthetic shapes to generate intermediate shapes to propagate correspondences, we collect these interpolated latent codes into an empirical distribution $p_{\sync}$ and define the regularization loss as
\begin{equation}
l_{\textup{reg, II}}(\phi) = \underset{\bs{z}\sim p_{\sync}}{\mathbb{E}} r(f^{\phi}(\cdot, \bs{z})).
\label{Eq:Regu:Loss:II:Stage:I}    
\end{equation}
We adopt the approach in GenCorres~\cite{DBLP:conf/iclr/0005HSBH24} to define $r(f^{\phi}(\cdot, \bs{z}))$ but make two significant modifications. First, we define $r(f^{\phi}(\cdot, \bs{z}))$ using a reduced deformation model, which greatly increases the efficiency of training while making the effectiveness of regularization more prominent. Specifically, consider a mesh discretization of $f^{\phi}(\cdot, \bs{z})$ with $m$ vertices $\bs{p}^{\phi}(\bs{z})\in \R^{3m}$. Consider $l \ll m$ deformation nodes obtained by farthest point sampling ($m = 2000$ and $l = 200$ in our experiments). With $\set{N}_i$ we collect adjacent vertices of node $i$, where adjacent $\set{N}_i$ overlap and $\{\set{N}_i\}$ cover all mesh vertices. With this setup, we parameterize $\bs{p}^{\phi}(\bs{z}+\epsilon\bs{v}) \approx  \bs{p}^{\phi}(\bs{z}) + \epsilon \bs{d}(\bs{z},\bs{v})$ in the neighborhood of $\bs{z}$ where $\bs{d}(\bs{z},\bs{v}):=$ 
\begin{align}
\underset{\bs{d}}{\textup{argmin}} & \min_{s_i,\bs{c}_i}\sum\limits_{i=1}^{l}\big(\sum\limits_{j\in \set{N}_i} \|A_i(\bs{p}_i^{\phi}(\bs{z})-\bs{p}_j^{\phi}(\bs{z})) - (\bs{d}_i-\bs{d}_j)\|^2 + \eta s_i^2\big)\nonumber \\
s.t. & \quad \frac{\partial f^{\phi}}{\partial \bs{x}}(\bs{p}_j^{\phi}(\bs{z}),\bs{z})\bs{d}_j + \frac{\partial f^{\phi}}{\partial \bs{z}}(\bs{p}_j^{\phi}(\bs{z}),\bs{z})\bs{v} = 0, \quad 1\leq j \leq m
\label{Eq:Cons:Opt}
\end{align}
where 
$$
A_i = s_i I_3 + \bs{c}_i\times I_3.
$$
Through some derivations (see the supp. material), we can rewrite Eq.~(\ref{Eq:Cons:Opt}) as
\begin{equation}
\bs{d}^{\bs{v}}:= \underset{\bs{d}}{\textup{argmin}} \ \ \bs{d}^T L^{\phi}(\bs{z}) \bs{d} \quad s.t. \quad C^{\phi}(\bs{z}) \bs{d}+ F^{\phi}(\bs{z}) \bs{v} = 0.
\label{Eq:Cons:Opt2}
\end{equation}
This leads to an explicit expression of 
$$
d^{\bs{v}} = G^{\phi}(\bs{z}) \bs{v},\quad G^{\phi}(\bs{z}) = -(I,0)\left(
\begin{array}{cc}
L^{\phi}(\bs{z}) & {C^{\phi}(\bs{z})}^T \\
C^{\phi}(\bs{z}) & 0
\end{array}
\right)^{\dagger}\left(
\begin{array}{c}
0 \\
F^{\phi}(\bs{z})
\end{array}
\right)\bs{v}. 
$$
Second, unlike merely minimizing ${\bs{d}^{\bs{v}}}^TL^{\phi}(\bs{z})\bs{d}^{\bs{v}}$, we define the regularization loss at $\bs{z}$ by integrating the ratio $\frac{{\bs{d}^{\bs{v}}}^TL^{\phi}(\bs{z})\bs{d}^{\bs{v}}}{{\bs{d}^{\bs{v}}}^T\bs{d}^{\bs{v}}}$, which measures the relative deformation with respect to the absolute displacement. This becomes important as the distribution of shapes in the AE latent space becomes irregular. Specifically, we have 
\begin{equation}
r(f^{\phi}(\cdot, \bs{z}) = \int_{\bs{v}\in \set{S}^d}\frac{{\bs{d}^{\bs{v}}}^TL^{\phi}(\bs{z})\bs{d}^{\bs{v}}}{\|\bs{d}^{\bs{v}}\|^2}d\bs{v}.    
\label{Eq:Def:Regu}
\end{equation}
Combing Eqs.(\ref{Eq:Data:Loss:Stage:II}), (\ref{Eq:Regu:Loss:I:Stage:I}), and (\ref{Eq:Regu:Loss:II:Stage:I}), the total training loss for stage II is 
\begin{equation}
\min\limits_{\gamma, \phi} l_{\data}(\phi, \gamma) + \lambda l_{\reg, I}(\gamma) + \mu l_{\reg, II}(\phi).
\label{Eq:Stage:I:Total:Loss}
\end{equation}

To ensure that the latent code of $S_0$ is at the origin of the latent space, we sample $S_0$ once every batch and penalize the $L^2$-norm of its predicted latent.

\subsection{Stage III: Shape Generator Initialization}
\label{sec:method_stage3}

\begin{figure}[t]
\includegraphics[width=\textwidth]{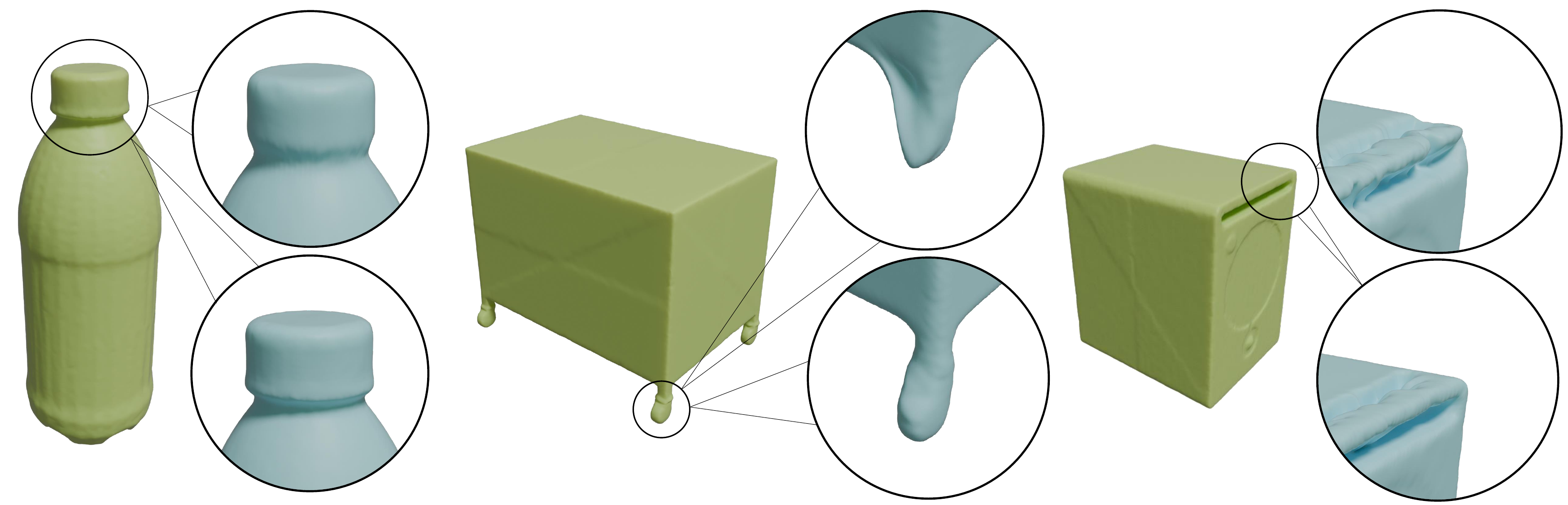}
\caption{(Top) If we deform $S_0$ to align $S_i$ through intermediate shapes defined by $f^{\phi}(\cdot, \bs{z})$ the alignment is poor, as some intermediate shapes have large distortions. (Bottom) When we deform $S_0$ to align $S_i$ through intermediate shapes between adjacent training shapes on a spanning tree rooted at $S_0$, the alignment is more accurate as intermediate shapes are much more meaningful.}
\label{Figure:Corres:Initialization}  
\vspace{-0.2in}  
\end{figure}

The goal of the third stage is to compute the initial dense correspondences between $S_0$ and each training shape $S_i$, which will be used to initialize the shape generator $\bs{g}^{\theta}$. A naive approach is to use $f^{\phi}(\cdot,\bs{z})$ to generate intermediate shapes between $S_0$ and $S_i$ and to align $S_0$ with these intermediate shapes sequentially to obtain the correspondences with $S_i$. However, this approach does not work when $S_i$ and $S_0$ are very different from each other, as many intermediate shapes present large deformations (see Figure~\ref{Figure:Corres:Initialization}(Top)).

Our idea is to introduce a more meaningful interpolation path from $S_0$ to $S_i$ using the training data. Specifically, we build a minimum spanning tree among $\set{S}$ where the root is $S_0$. The weight of each edge $(S_i, S_j)$ is given by the distance $\|e^{\gamma}(S_i)-e^{\gamma}(S_j)\|$ between $S_i$ and $S_j$ in the latent space. We also use $f^{\phi}(\cdot, \bs{z})$ to generate intermediate shapes between $S_i$ and $S_j$ to propagate the correspondences between $S_0$ and $S_i$ to $S_j$. Correspondence propagation employs the standard non-rigid registration approach with an as-rigid-as possible deformation prior~\cite{DBLP:journals/tvcg/TamCLLLMMSR13}. As shown in Figure~\ref{Figure:Corres:Initialization}(Bottom), this approach yields accurate correspondences for shapes $S_i$ that present large variations from $S_0$.

Let $\bs{p}^i\in \R^{3n}$ be the correspondences of the mesh vertices $\bs{c}\in \R^{3n}$ of $S_0$ on $S_i$. We initialize the mesh generator $\bs{g}^{\theta}(\bs{z},\cdot)$ using the standard $L^2$ loss:
\begin{equation}
\min\limits_{\theta} \sum\limits_{i=1}^{N}\sum\limits_{j=1}^{n}\|\bs{g}^{\theta}(e^{\gamma}(S_i),\bs{c}_j)-\bs{p}_j^i\|^2.
\label{Eq:StageII:Loss}    
\end{equation}

\subsection{Stage IV: Shape Generator Fine-tuning}

\begin{figure}[t]
\includegraphics[width=\textwidth]{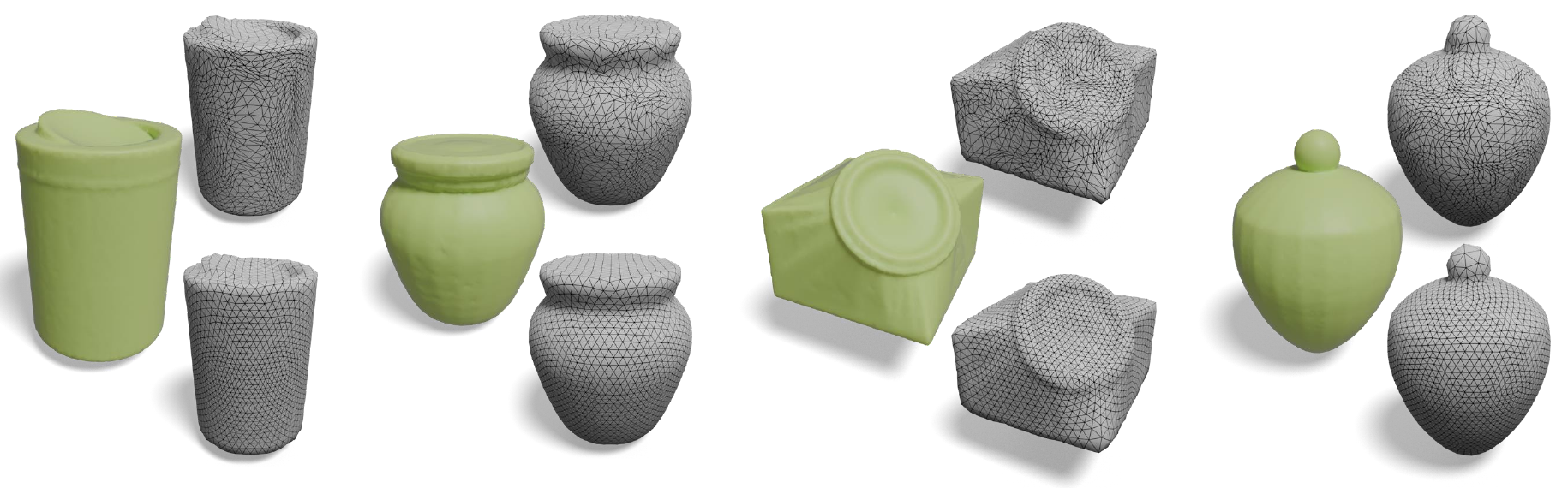}
\caption{Comparisons between (Top) initial correspondences between $S_0$ and each training shape and (Bottom) the correspondences obtained by optimized shape generator. We can see that the optimized correspondences are smoother and align the training data (green) better than the initial correspondences. This is attributed to the deformation loss on the mesh generator, which optimizes all correspondences together. }
\label{Figure:Comparisons}   
\end{figure}

The fourth stage jointly refines the shape generator $\bs{g}^{\theta}$ and the encoder $e^{\gamma}$. The training loss consists of a modified data loss and a deformation regularization loss on $\bs{g}^{\theta}$. Its effect on improving the quality is shown in Figure~\ref{Figure:Comparisons}.

The modified data loss is given by
\begin{equation}
l_{\textup{data}}(\gamma, \theta) := \sum\limits_{i=0}^{N} \int_{\bs{p}\in \set{S}^2} d^2\big(\bs{g}^{\theta}(e^{\gamma}(S_i),\bs{p}), S_i\big)
\label{Eq:Data:Loss:Stage:IV}    
\end{equation}
where $d^2\big(\bs{q}, S_i)$ is the squared distance between $\bs{q}$ and the triangular mesh $S_i$.


The deformation regularization loss models deformation priors on $\bs{g}^{\theta}(\bs{z},\bs{p})$. As $\bs{g}^{\theta}(\bs{z},\bs{p})$ is continuous in $\bs{p}$, we model the deformation regularization loss based on the Jacobian $J_{\bs{p}}^{\theta}(\bs{z},\bs{p})= \frac{\partial \bs{g}^{\theta}(\bs{z},\bs{p})}{\partial \bs{p}} \in \R^{3\times 2}$ which is computed in a local tangent frame at each sphere point $p$, formed by two orthonormal vectors obtained by Gram-Schmidt against the radial direction at $p$. The Jacobian itself is computed via PyTorch \texttt{autograd}. It is easy to see that the desired properties for $\bs{p}^{\theta}(\bs{z},\bs{p})$ to be an isometric and conformal mapping are that ${J_{\bs{p}}^{\theta}(\bs{z},\bs{p})}^T J_{\bs{p}}^{\theta}(\bs{z},\bs{p})$ is the identity matrix $I_2$ and the scaled identity matrix, respectively. Therefore, we define the following regularization loss 
\begin{equation}
l_{\textup{def}}(\theta) = \underset{\bs{z}\sim p_{\sync},\bs{p}\in \set{S}^2}{\mathbb{E}} r({J_{\bs{p}}^{\theta}(\bs{z},\bs{p})}^T J_{\bs{p}}^{\theta}(\bs{z},\bs{p})) 
\label{}
\end{equation}
where
$$
r(A) = \beta \|A-I_2\|^2 + \|A-\frac{\langle A, I_2\rangle}{2} I_2\|^2. 
$$
We set $\beta = 0.01$ in all of our experiments. 

The training loss at this stage is then given by
$$
\min\limits_{\theta}  l_{\textup{data}}(\gamma, \theta)+ \overline{\mu}l_{\textup{def}}(\theta).
$$

\section{Experimental Results}

\subsection{Experimental Setup}

We collected a dataset of genus-0 shapes from the ShapeNet \cite{chang2015shapenet} dataset, totaling 1K shapes. These shapes cover diverse objects. We use 900 shapes for training and 100 shapes for testing. For baseline comparisons, we choose five state-of-the-art single-shape spherical parameterization approaches, i.e., CMCF~\cite{DBLP:journals/cgf/KazhdanSB12} and SMAT ~\cite{DBLP:journals/cgf/SchmidtPK23}, AHSP~\cite{Hu-2017-AHSP}, ARAP~\cite{10.1016/j.gmod.2014.03.016}, VC25~\cite{DBLP:journals/vc/ChenXC25}. Here, CMCF is flow-based, while the other four baselines are optimization-based. 

We consider four metrics for experimental evaluation. Similar to VC25, the first three metrics evaluate the quality of each spherical parameterization via its geometric distortion. Three of these metrics consider the distortions of each triangle $t$ and each method. Specifically, we compute the Jacobian $J_t$ between the image of $t$ on the sphere and itself. We report two distortions based on $J_t$:
$$
\sigma_t = \max\big(\sigma_{\max}(J_t), \frac{1}{\sigma_{\min}(J_t)}\big), \ \ \sigma_t^{\textup{con}} = \max\big(\frac{\sigma_{\max}(J_t)}{\sigma_{\mean}(J_t)}, \frac{\sigma_{\mean}(J_t)}{\sigma_{\min}(J_t)}\big).
$$
where $\sigma_{\max}(J_t)$, $\sigma_{\min}(J_t)$, and $\sigma_{\mean}(J_t)$ are the maximum, minimum, and mean singular values of $J_t$. We report the mean and maximum of $\sigma_t$ and the maximum of $\sigma_t^{\textup{con}}$ among all triangles. 
The fourth metric evaluates the consistency of the spherical parameterizations. First we compute a graph $\set{G} = (\{S_1,\cdots, S_N\},\set{E})$ that connects similar input shapes. $\set{E}$ connects each shape $S_i$ to its $6$ nearest neighbors according to the Hausdorff distance. Let $m_{ij}: S_i\rightarrow S_j$ be the ground-truth mapping between two similar shapes $(S_i,S_j)\in \set{E}$. 
We use ground truth mappings $m_{ij}$ provided by KeypointNet\cite{you2020keypointnet}.
We compute the differences $\|m_{ij}\circ m_i^{-1}(S_0), m_j^{-1}(S_0)\|$ between images of the same sample points of $S_0$ on $S_i$ and $S_j$. We report the average of $c(m_i,m_j) = d(m_{ij}\circ m_i^{-1}(S_0), m_j^{-1}(S_0))/d_{\textup{Haus}}(S_i, S_j)$.  The lower the value of $c(m_i,m_j)$ the more consistent $m_i$ and $m_j$. Note that for all baseline approaches, we employ M{\"{o}}bius registration to factor out the underlying M{\"{o}}bius group ~\cite{DBLP:journals/cgf/BadenCK18} before calculating $c(m_i, m_j)$. 


\subsection{Sphere Parameterization Quality}

\begin{table}
\centering
\begin{tabular}{c|ccc|c|c}
&$\mean(\sigma_t)$ & $\max(\sigma_t)$& $\mean(\sigma_t^{\textup{con}})$ & $\mean(c(m_i,m_j))$ & run-time\\\hline
CMCF& 24.22 & \num{1.13e3} & 1.029 & 1.626 &$\sim 5$ min\\
ARAP& 177.19 & \num{1.219e6} & 1.029 & 1.608 &$\sim 6$ min\\
AHSP& 1.578 & 2.349 & 1.232 & 6.006 &$\sim 32$ sec\\
SMAT& 1.503 & 3.392 & 1.049 & 5.122 &$\sim 2$ min\\
VC25& 12.520 & 216.70 & 6.892 & 6.573 &$\sim 5$ min\\\hline
GenSP& 1.681 & 7.719 & 1.212 & 1.425 & $<1$ sec\\\hline
No-DA & 2.326 & 614.14 & 1.443 & 3.164 &- \\
No-GP & 1.866 & 23.228 & 1.214 & 1.613 &- \\
No-GR & 2.296 & 271.26 & 1.627 & 1.740 & - \\\hline
\end{tabular}
\vspace{7px}
\caption{(Top) Baseline comparisons on the held-out test set between GenSP and five state-of-the-art spherical parameterization approaches, CMCF~\cite{DBLP:journals/cgf/KazhdanSB12}, SMAT ~\cite{DBLP:journals/cgf/SchmidtPK23}, AHSP~\cite{Hu-2017-AHSP}, ARAP~\cite{10.1016/j.gmod.2014.03.016}, and VC25~\cite{DBLP:journals/vc/ChenXC25}. We report mean and max of isometric distortion --- $\mean(\sigma_t)$ and $\max(\sigma_t)$, max of conformal distortion --- $\max(\sigma_t^{\textup{con}})$, and the consistency of spherical parameterizations --- $\mean(c(m_i,m_j))$. (Bottom) Ablation study results. (No-DA) No data augmentation. (No-GP). Matching the sphere to each shape directly using intermediate shapes from the implicit generator, i.e., without using a path of training shapes. (No-GR) Without using regularization losses when training the generative models.}
\vspace{-0.3in}
\label{Table:Quan:Eval}
\end{table}

Table~\ref{Table:Quan:Eval} presents quantitative comparisons between GenSP and the baseline approaches. We can see that GenSP outperforms all baseline approaches considerably across all metrics. Compared to CMCF, GenSP reduces $\mean(\sigma_t)$ and $\max(\sigma_t)$ by $14.3\times$ and $146\times$ respectively. Even for $\mean(\sigma_t^{\textup{con}})$ which favors CMCF, GenSP only increases by $1.17\times$. 
GenSP's $\mean(\sigma_t)$, $\max(\sigma_t)$, and $\mean(\sigma_t^{\textup{con}})$ also remain on par with per-shape optimization based methods. 

With respect to the consistency metric, $\mean(c(m_i,m_j))$, GenSP leads to significant performance gains compared to per-shape optimization methods, i.e., $3.6\times$ better than the top-performing optimization-based baseline SMAT. It also outperforms flow based approaches to a lesser extent. This is expected as GenSP explicitly models the consistency of spherical parameterizations as an optimization objective. 

\begin{figure}[h]
\centering
\includegraphics[width=0.9\textwidth]{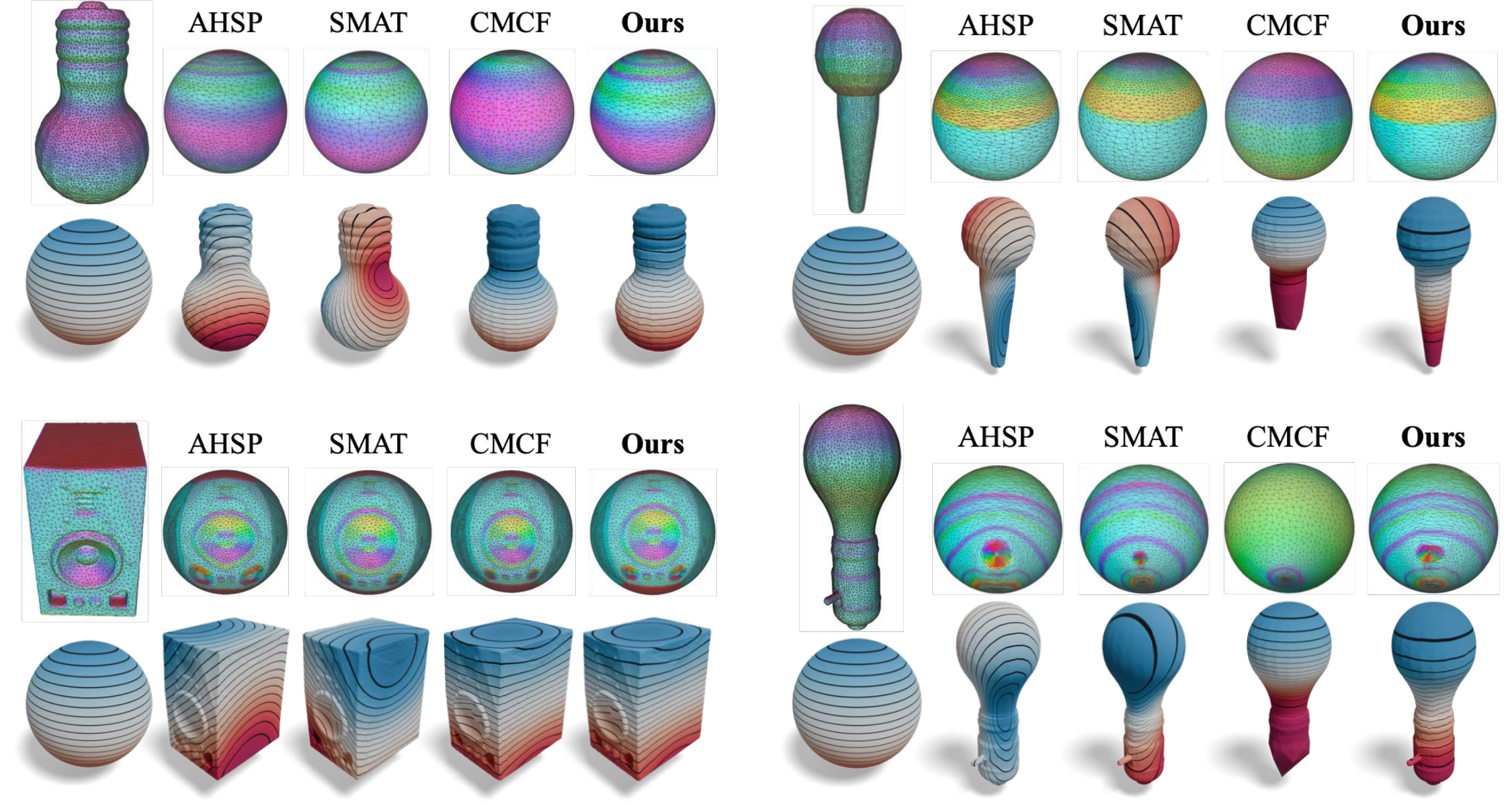}
\caption{Additional qualitative spherical parameterization results. The top row in each block visualizes the mapping of each method from a shape to the sphere. (The spheres are manually rotated for ease of visualization.) The bottom row shows the mapping of each method from the sphere to the shape. We can see that our method offers more consistent and high-quality parameterizations.}
\label{Fig:Qual:Eval}    
\end{figure}

Figure~\ref{Fig:Qual:Eval} shows qualitative comparisons between GenSP and baseline approaches. Qualitative results are consistent with quantitative results. The individual spherical parameterizations of GenSP exhibit fewer distortions, which is expected. Moreover, GenSP is the only approach that offers consistent parameterizations across the shape collection, in which semantic feature keypoints are in correspondence. 

\begin{figure}[h]
\begin{minipage}{0.55\textwidth}
  \scriptsize
  \setlength{\tabcolsep}{2pt}
  \begin{tabular}{>{\centering\arraybackslash}p{0.18\textwidth}|>{\centering\arraybackslash}p{0.18\textwidth}>{\centering\arraybackslash}p{0.18\textwidth}>{\centering\arraybackslash}p{0.18\textwidth}>{\centering\arraybackslash}p{0.18\textwidth}}
    \hline
    Source     & GenSP  & Diff3F & Diffumatch & ULRSSM\\ \hline
    \includegraphics[height=0.16\textwidth]{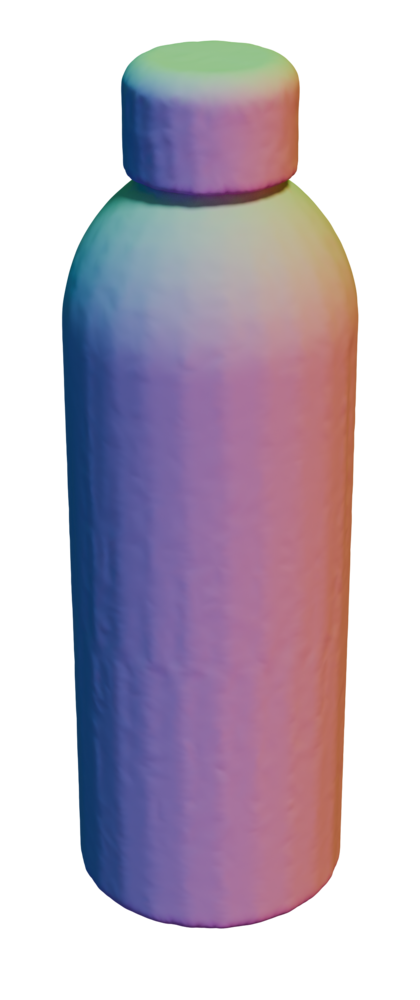}     &  \includegraphics[height=0.16\textwidth]{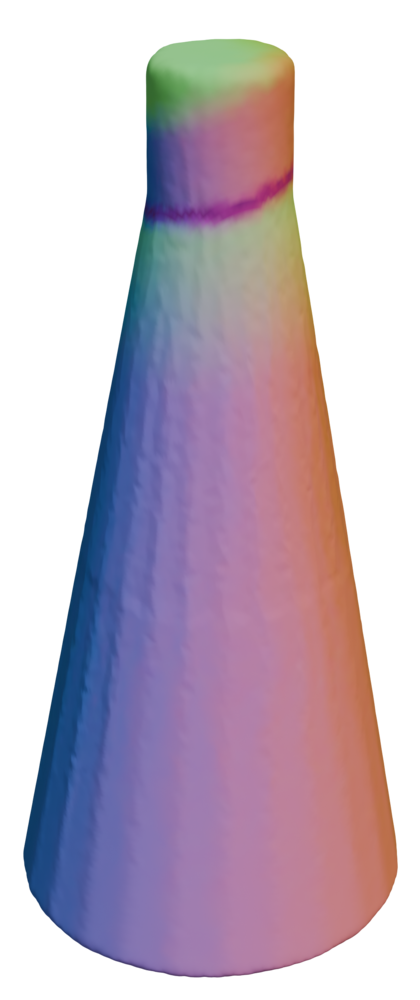}&
    \includegraphics[height=0.16\textwidth]{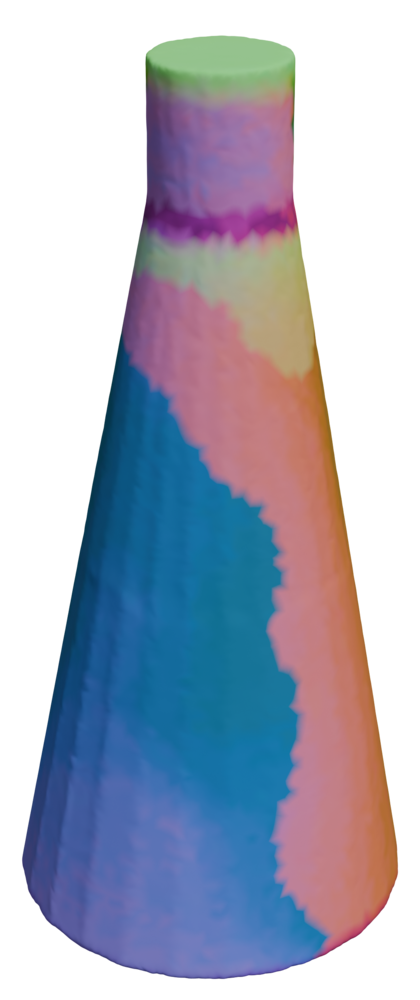}& \includegraphics[height=0.16\textwidth]{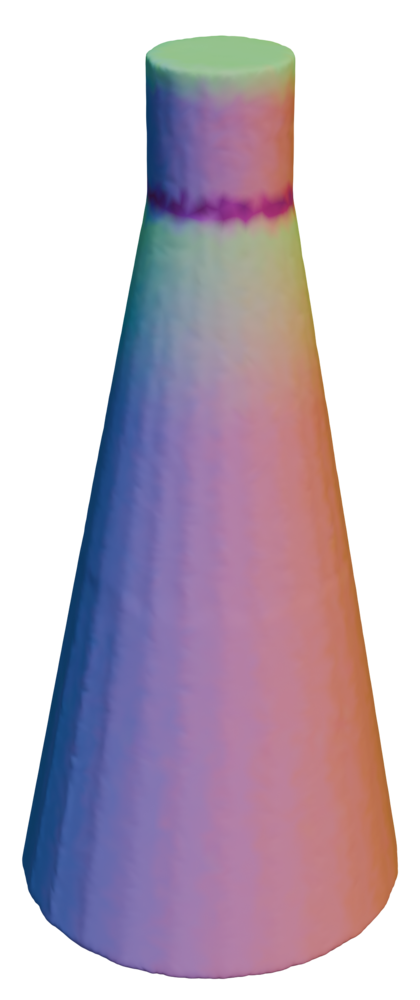}&
    \includegraphics[height=0.16\textwidth]{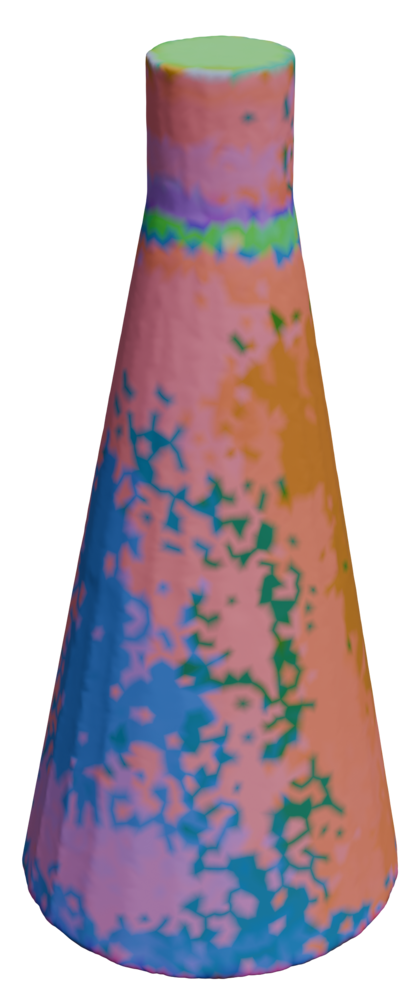}\\
    \includegraphics[height=0.16\textwidth]{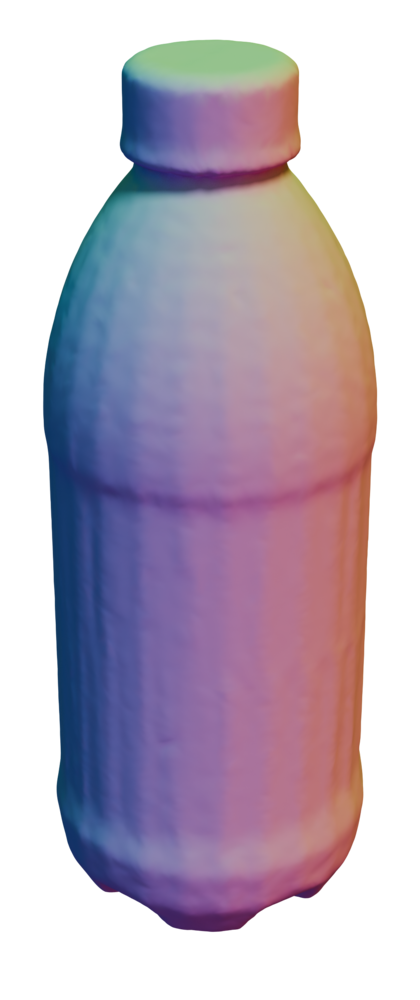}&
    \includegraphics[height=0.16\textwidth]{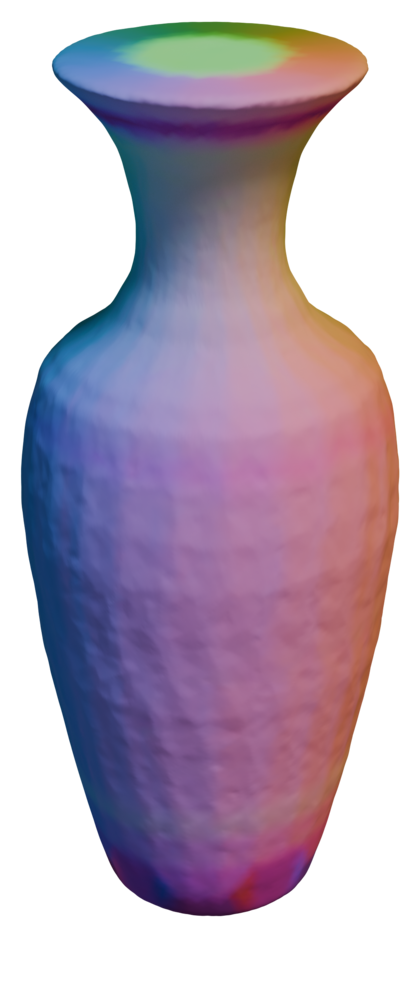} &
    \includegraphics[height=0.16\textwidth]{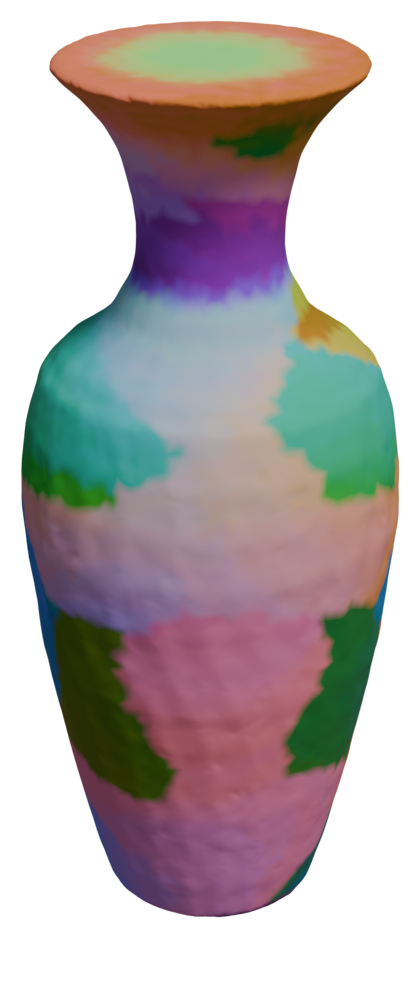} &
    \includegraphics[height=0.16\textwidth]{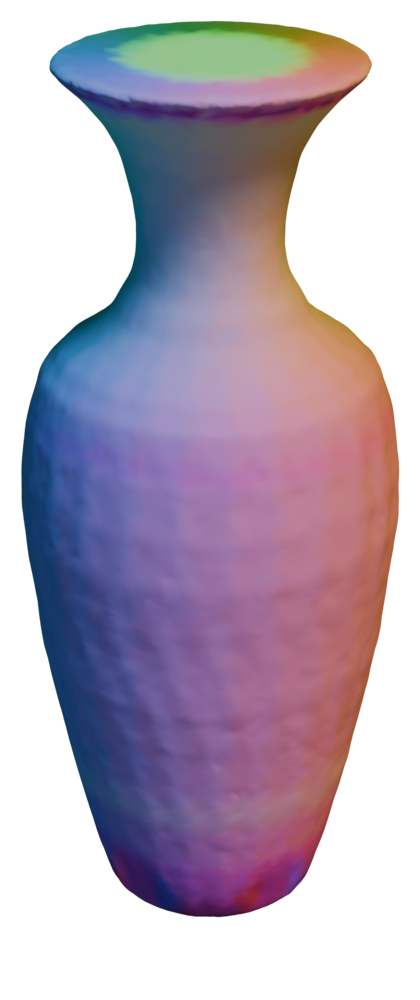}&
    \includegraphics[height=0.16\textwidth]{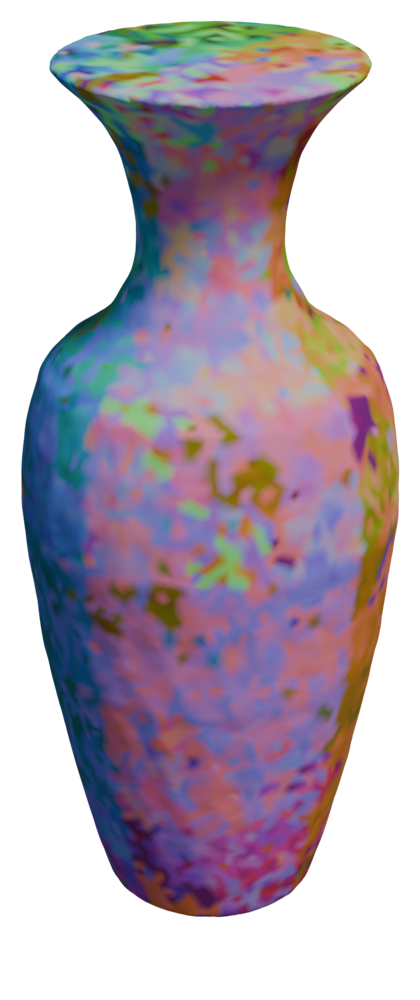} \\
    \hline
    Time & $\sim0.3$s & $\sim9$ min & $\sim80$s & $\sim50$s
  \end{tabular}
  \caption{Comparison against modern correspondence methods.}
  \label{Fig:Modern:Shape:Matching}
\end{minipage}
\hfill
\begin{minipage}{0.40\textwidth}
  \fontsize{7.5pt}{9pt}\selectfont 
  \setlength{\tabcolsep}{3pt}
  \centering
  \begin{tabular}{l|c|c}
    \hline
    Method & Bad  & \# Affected \\
    & face \% & meshes \\ \hline
    SMAT & 0.004 & 4 \\\hline
    AHSP & 0.002 & 3 \\\hline
    CMCF & 0.241 & 15 \\\hline
    ARAP & 0.434 & 31 \\\hline
    VC25 & 1.898 & 91 \\\hline
    GenSP & 0.099 & 9 \\
    \hline
  \end{tabular}
  \captionof{table}{Percentage of self-intersecting and degenerate faces in spherical parameterizations.}
  \label{tab:bijectivity}
\end{minipage}
\end{figure}

\textbf{Bijectivity Guarantees:} GenSP does not provide formal bijectivity guarantees, but it is empirically near-bijective. On our 100-shape ShapeNet test-set, 91 shapes produce fully bijective parameterizations and the mean per-shape rate of self-intersecting or degenerate faces is 0.099\% as shown in Table ~\ref{tab:bijectivity}. This is comparable to optimization-based methods that explicitly target bijectivity per shape and is substantially better than other flow-based or stereographic methods.

\subsection{Ablation Study}
\label{Sec:Ablation}

\subsubsection{No data augmentation} The first ablation study removes the data augmentation step and uses only input shapes to learn the generative models. As shown in Table ~\ref{Table:Quan:Eval}, all metrics of this variant drop significantly. This is because extrinsic generative models fail to recover the underlying complex deformations between the sphere and these shapes without training on shapes that bridge them. 

\subsubsection{No intermediate training shapes} The second ablation study replaces correspondence propagation along a path of adjacent training shapes in stage III with intermediate shapes defined by the implicit shape generator via straight-line interpolation in the latent space. Table~\ref{Table:Quan:Eval} shows that this variant leads to notable performance drops in all metrics, except the consistency metric, which is similar to CMCF.  This is because the mesh generator depends on the quality of the initial correspondences and cannot recover from severe errors in initial correspondences, which leads to poor isometry errors. On the other hand, as the latent space is still organized according to the intermediate CMCF shapes, the consistency metric remains similar to that of CMCF.

\subsubsection{No geometric regularizations} The third ablation study drops the geometric regularizations used in training the implicit and mesh generators. Table~\ref{Table:Quan:Eval} shows that this variant leads to noticeable performance drops. This is expected as geometric regularizations are critical for determining how to interpolate shapes between sparse shape samples, e.g., between complex shapes. Figure \ref{Figure:Comparisons} depicts the improvement in the quality of parameterizations when this regularization is used, which also contributes to the improvement in metrics.

\subsection{Applications}

In Figure \ref{Fig:Modern:Shape:Matching}, we compare GenSP against three recent correspondence methods (Diff3F~\cite{diff3d}, Diffumatch~\cite{diffumatch}, and ULRSSM~\cite{ulrssm}). GenSP produces more coherent inter-class and intra-class surface correspondences while running at least $100\times$ faster. ULRSSM fails on axially symmetric shapes due to rotationally non-unique features. We show a further application in shape morphing and comparison with classical methods in Sec. \ref{sec:additional_results}.

\section{Conclusions and Future Work}

In this paper, we have presented a data-driven approach that computes a consistent spherical parameterization among a collection of genus-0 shapes. We have shown how to formulate this problem as learning a neural deformation-based mesh generative model. We present a technique to compute consistent correspondences between training shapes and the sphere to initialize the mesh generative model. The experimental results reveal that GenSP retains the  high-quality isometric factors  yielded by per-shape optimization based spherical parameterization approaches while simultaneously providing consistent parameterizations like flow based methods. 

There are ample opportunities for future research. First is how to extend this framework to higher genera shapes. CMCF could be replaced with a curvature flow admitting non-constant final mean curvature (e.g., Willmore flow variants \cite{Crane:2013:RFC}) and the sphere with high-genera templates.
Second, the shape generative models are extrinsic, while spherical parameterization is an intrinsic problem. Developing an intrinsic shape generative model for spherical parameterization is another interesting problem.
\section{Acknowledgments}

This project was supported by NSF-2047677, 2413161, 2504906, and 2515626; GIFTs from Adobe, Google, and Ericsson; and computing support on the Vista GPU Cluster through the Center for Generative AI (CGAI) and the Texas Advanced Computing Center (TACC) at UT Austin.
%
%
\bibliographystyle{splncs04}
\bibliography{neural}
\clearpage
\appendix

\section{Derivation of Eq.~\ref{Eq:Cons:Opt}}

This section presents a matrix reformulation of Eq.~\ref{Eq:Cons:Opt}, which is 
\begin{equation}
\bs{d}^T L^{\phi}(\bs{z})\bs{d}  = \min_{s_i,\bs{c}_i}\sum\limits_{i=1}^{l}\big(\sum\limits_{j\in \set{N}_i} \|A_i(\bs{p}_i^{\phi}(\bs{z})-\bs{p}_j^{\phi}(\bs{z})) - (\bs{d}_i-\bs{d}_j)\|^2 + \eta s_i^2\big)
\label{Eq:Cons:Opt2}
\end{equation}
where 
$$
A_i = s_i I_3 + \bs{c}_i\times I_3.
$$
Denote $\bs{x}\in \R^{4l}$ where $\bs{x}_i = (s_i;\bs{c}_i)$. Introduce $L\in \R^{m\times m}$, $B^{\phi}(\bs{z})\in \R^{3m\times 4l}$, and $H^{\phi}(\bs{z})\in \R^{4l\times 4l}$ where
\begin{align*}
L_{ii} &= \left\{
\begin{array}{cc}
|\set{N}_i|& i= j, \quad 1\leq i \leq l \\
|\{i'|i\in \set{N}_{i'}\}| & \quad i=j, \quad l+1\leq i\leq m\\
-1 & j\in \set{N}_i, \quad \textup{or} i\in \set{N}_j \\
0 & \textup{otherwise}
\end{array}
\right.\,\\
B_{ji}^{\phi}(\bs{z}) & = \left\{
\begin{array}{cc}
\big(\bs{q}_{ij}^{\phi}(\bs{z}),-\bs{q}_{ij}^{\phi}(\bs{z})\times \big)& i = i\nonumber \\
-\big(\bs{q}_{ij}^{\phi}(\bs{z}),-\bs{q}_{ij}^{\phi}(\bs{z})\times \big)& j\in \set{N}_i\nonumber \\
0 & \textup{otherwise}
\end{array}
\right.\,\\
H_{ij}^{\phi}(\bs{z}) & = \left\{
\begin{array}{cc}
\sum\limits_{j'\in \set{N}_i}\left(
\begin{array}{cc}
\|\bs{q}_{ij'}^{\phi}(\bs{z})\|^2 + \eta &0\\
0 & \|\bs{q}_{ij'}^{\phi}(\bs{z})\|^2I_3- \bs{q}_{ij'}^{\phi}(\bs{z}){\bs{q}_{ij'}^{\phi}(\bs{z})}^T
\end{array}
\right) & i = j \\
 & \textup{otherwise}
\end{array}
\right.\,\\
\end{align*}
where
$$
\bs{q}_{ij}^{\phi}(\bs{z}) = \bs{p}_i^{\phi}(\bs{z})-\bs{p}_j^{\phi}(\bs{z}).
$$
Then 
\begin{align}
& \quad \bs{d}^T L^{\phi}(\bs{z})\bs{d} \nonumber \\
=& \quad\min\limits_{\bs{x}}\quad\Big(\bs{d}^T(L\otimes I_3))\bs{d} -2\bs{d}^T B^{\phi}(\bs{z}) \bs{x} + \bs{x}^T H^{\phi}(\bs{z})\bs{x}\Big) \nonumber \\
=&\quad \bs{d}^T\Big(L\otimes I_3-B^{\phi}(\bs{z})H^{\phi}(\bs{z}){B^{\phi}(\bs{z})}^{-1}\Big)\bs{d}, \quad \Big(\bs{x} = {H^{\phi}(\bs{z})}^{-1}(B^{\phi}(\bs{z})^T\bs{d})\Big) \quad 
\end{align}
meaning
$$
L^{\phi}(\bs{z}) = L\otimes I_3-B^{\phi}(\bs{z})H^{\phi}(\bs{z}){B^{\phi}(\bs{z})}^{-1}.
$$

\section{Implementation Details}

\subsection{Latent Space Regularization}

A desirable property of the latent space is that similar shapes are reconstructed similarly and therefore assigned nearby latent codes. We find that the L1 penalty together with reconstruction loss encourages this desired behavior.  We verified that latent distances strongly correlate with 3D Hausdorff distances (Pearson’s $r = 0.723$).

\subsection{Registration}

As mentioned in Sect.~\ref{sec:method_stage3}, we construct paths from a template shape $S_0$ (the sphere) to the target shape $S_i$ via various intermediate shapes. Our goal is to build correspondences between the end points, and to do this we build correspondences between adjacent nodes in the path. Given two meshes in the path, $S_i$ and $S_j$, we first take their latents and interpolate them to get $n$ interpolated latents. We then decode them to get $n$ intermediate meshes between the two nodes in the path, say $(S_{i\rightarrow j})_1,\dots, (S_{i\rightarrow j})_n$. In our implementation, $n=5$. We then iteratively register $(S_{i\rightarrow j})_{n-1}$ to $(S_{i\rightarrow j})_{n}$ via standard non-rigid registration with an as-rigid-as possible deformation prior.
This gives a remeshed verion of $S_j$ with the same connectivity as $S_i$.

\subsection{Stitching}
Our next task is to ``stitch'' these independently registered segments together such that we can project the connectivity of a target shape $S$ onto the surface of the sphere.
Consider two registered segments $S_{i\rightarrow j}$ which has shape $S_j$ remeshed to connectivity of $S_i$, and $S_{j\rightarrow k}$ which has $S_k$ remeshed to connectivity of $S_j$.
Then essentially, as we have two differently remeshed versions of $S_j$, we can create a bijective map between them using barycentric interpolation of one vertex layout onto the surface of the other.
This map can be extended to $S_i$, which now has the same connectivity to $S_k$.
Iterating this process through all segments till the sphere, we can project the connectivity of $S$ to the sphere, creating the initializations that are used for training the mesh generative model.

\subsection{Dealing with Faulty Correspondences}
While our two-step registration and stitching process ensures that correspondences are propagated from the sphere to training shapes without severe self intersections, there are often $\sim5\%$ triangles that show self intersections on the sphere.
To ensure these faulty correspondences do not adverse affect the training of the mesh decoder, we simple omit samples from such regions from our mesh initiaization dataset.
Our mesh initialization training is robust enough to learn reasonable interpolations in these missing regions. 
The final joint-training stage relies on the ground truth shapes for minimizing the squared distance between the predicted shape and the original.
This final stage fixes the reconstruction in the under-sampled regions.

\subsection{Network Architecture}
\textbf{Encoder:} We use the PointNet++ architecture for the shape encoder module, which produces latents with 256 dimensions for any given shape.
\newline\textbf{Decoder:} Both our point decoder and mesh decoder follow the architecture of DeepSDF, and consist of 10 fully connected layers with 1024 hidden units in each of them. We have residual connections from the input to the third and sixth layer. In the case of the mesh decoder, we output a single SDF value, whereas for the mesh decoder produces a 3D location. We use 
\newline We train the implicit model for 600 epochs, the first 500 of which are without geometric regularization and the next 100 are including it. We train the mesh decoder for 2500 epochs, out of which the first 2000 are done without geometric regularization and the subsequent 500 include it.
For Stage II, $\lambda=0.1$ and $\mu=0.01$.
For Stage IV, $\bar{\mu} = 0.1$ and $\beta=0.01$.

\subsection{Training Details}
We train the implicit model on 8 $\times$ GH200 GPUS for 8 hours with a cumulative batch size of 96. Meanwhile, the mesh decoder training was done on a single H100 GPU with a batch size of 8. This takes a total of 14 hours wall-clock time. Learning rate for the implicit model training and mesh initialization is set at 1e-3, while the joint training stage uses a learning rate of 1e-4. All stages use cosine annealing scheduler and the Adam optimizer.

\section{More Results and Ablations}
\label{sec:additional_results}

We also trained and evaluated our method on the challenging human dataset D-FAUST (trained on 1180 shapes and tested on 120 held-out shapes). As shown in Table~\ref{tab:dfaust}, GenSP achieves the best consistency and mean conformal distortion by a high margin. 
Mean isometric distortion is on par with the optimization-based baselines, consistent with the ShapeNet results. 

\begin{table}[h]
\centering
\footnotesize
\begin{tabular}{l|cccc|c}
\hline
Metric & CMCF & ARAP & AHSP & SMAT & GenSP \\\hline
$\bar{\sigma}_t$ & 4.63e4 & 8.96e2 & \textbf{4.41} & 4.81 & 5.12 \\
$\max(\sigma_t)$ & 4.17e6 & 4.16e5 & \textbf{10.90} & 201.56 & 310.01 \\
$\bar{\sigma}_t^{\text{con}}$ & 1.71e5 & 8.95e4 & 3.95 & 2.88 & \textbf{2.33} \\\hline
$c(m_i, m_j)$ & 3.47 & 3.43 & 7.95 & 7.97 & \textbf{1.68} \\\hline
\end{tabular}
\vspace{0.3in}
\caption{Baseline comparisons on the held-out test set of D-FAUST (120 shapes) between GenSP and four state-of-the-art spherical parameterization approaches, CMCF~\cite{DBLP:journals/cgf/KazhdanSB12}, SMAT ~\cite{DBLP:journals/cgf/SchmidtPK23}. AHSP~\cite{Hu-2017-AHSP}, and ARAP~\cite{10.1016/j.gmod.2014.03.016}. VC25~\cite{DBLP:journals/vc/ChenXC25} fails to converge on D-FAUST.}
\label{tab:dfaust}
\end{table}

\begin{figure*}
\setlength{\tabcolsep}{1.5pt}
\begin{tabular}{cccccc}
\includegraphics[width=0.2\textwidth]{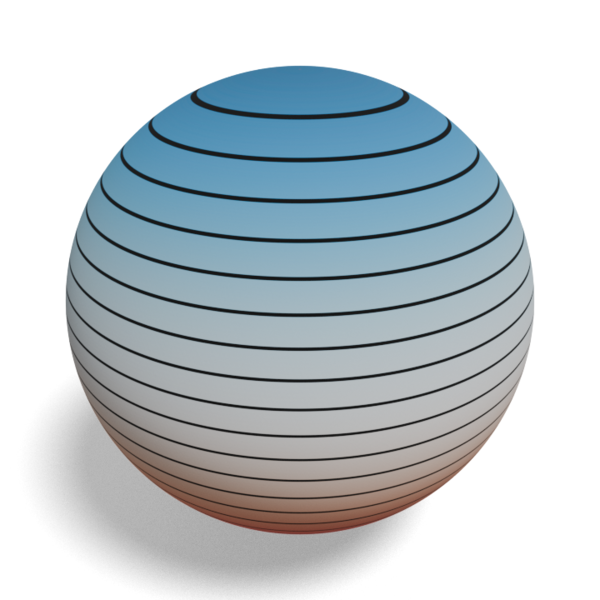}
& \centering{AHSP} & CMCF & SMAT & GenSP \\
\includegraphics[width=0.2\textwidth]{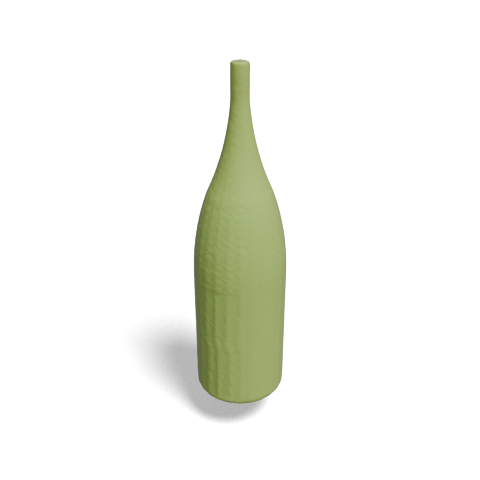} 
&\includegraphics[width=0.2\textwidth]{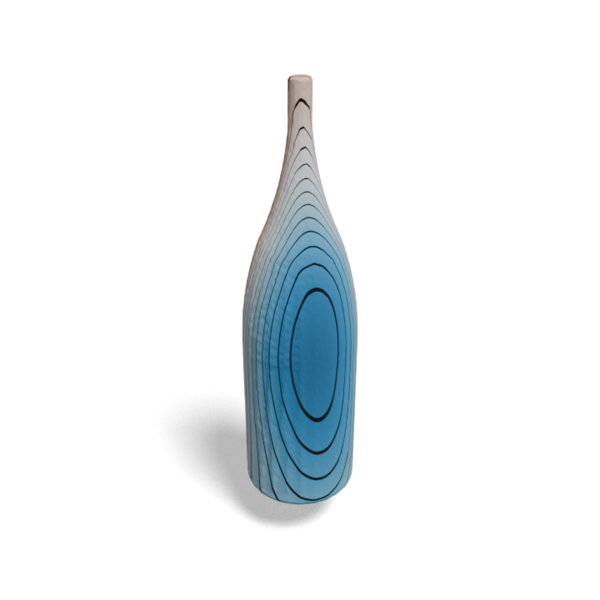} &\includegraphics[width=0.2\textwidth]{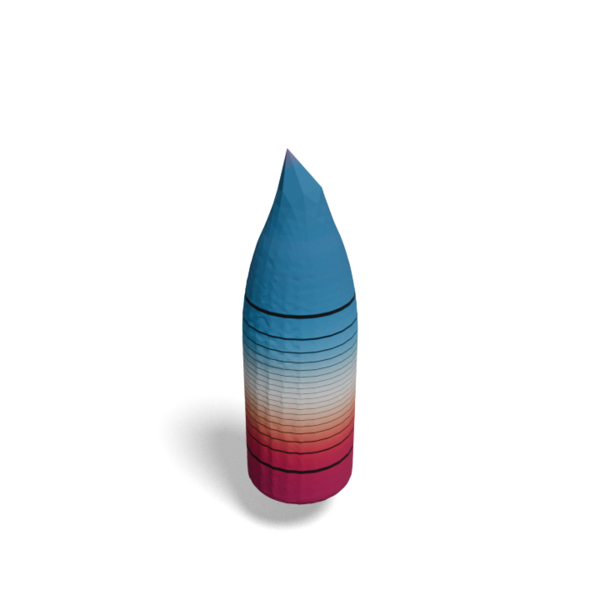}  &\includegraphics[width=0.2\textwidth]{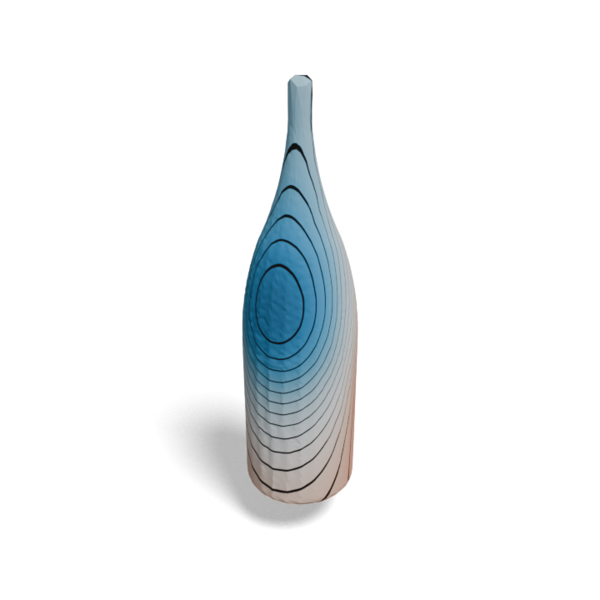} &\includegraphics[width=0.2\textwidth]{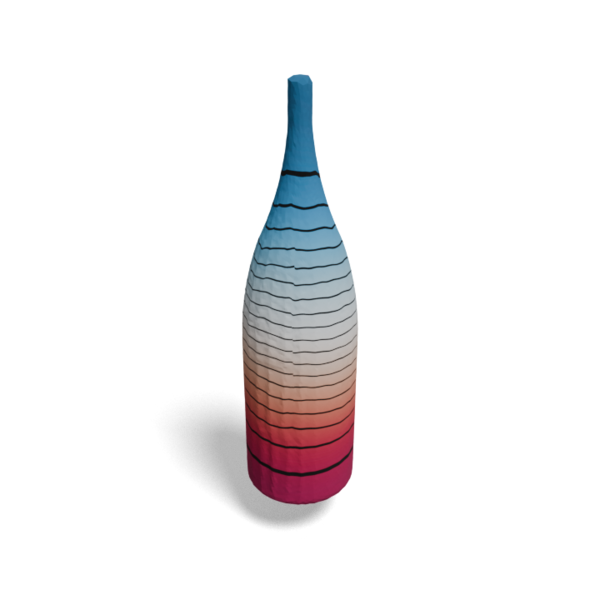} \\ 
\includegraphics[width=0.2\textwidth]{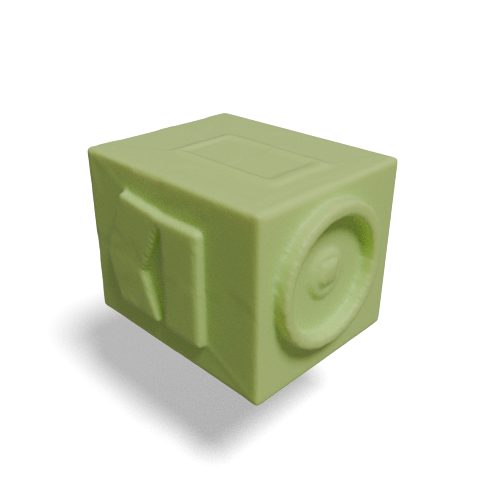} 
&\includegraphics[width=0.2\textwidth]{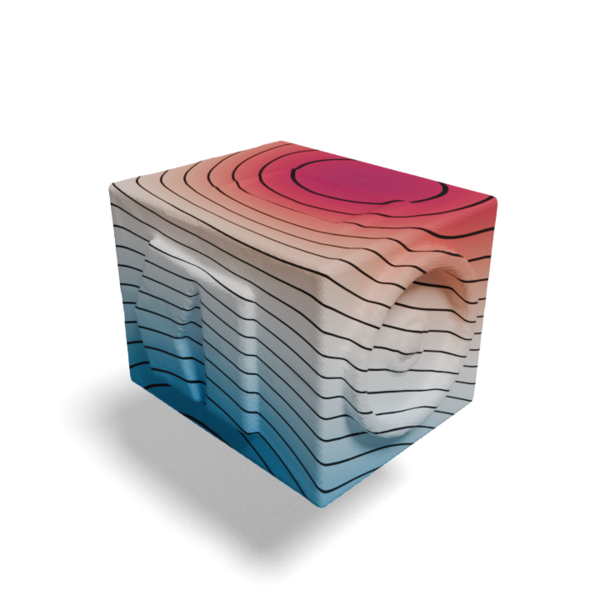}
&\includegraphics[width=0.2\textwidth]{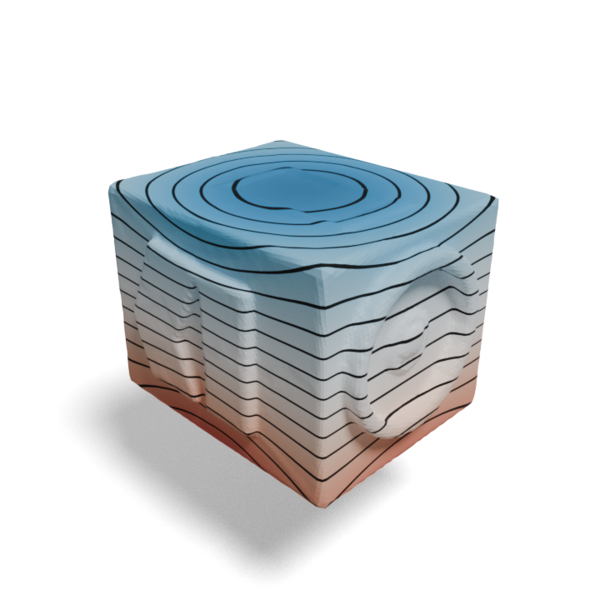} &\includegraphics[width=0.2\textwidth]{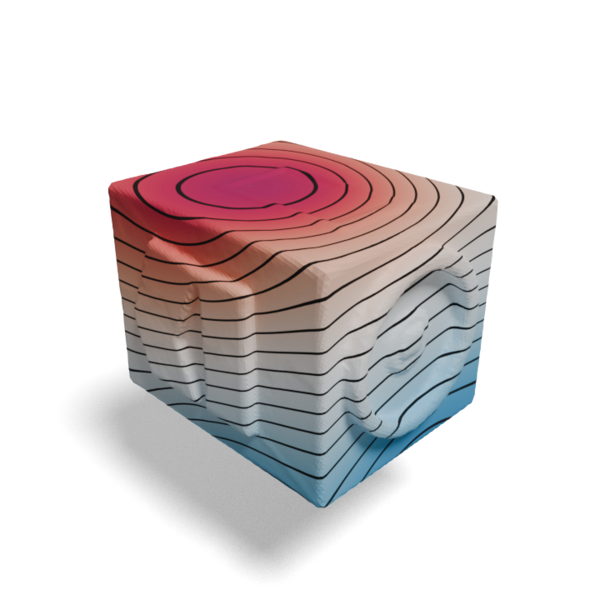} &\includegraphics[width=0.2\textwidth]{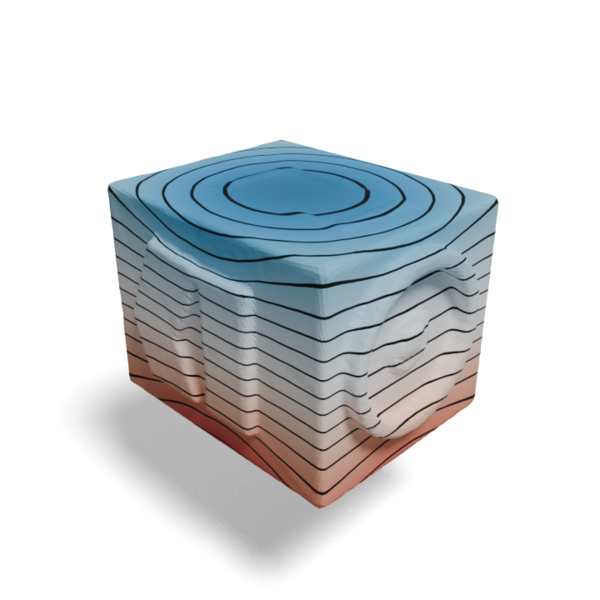} \\ 
\includegraphics[width=0.2\textwidth]{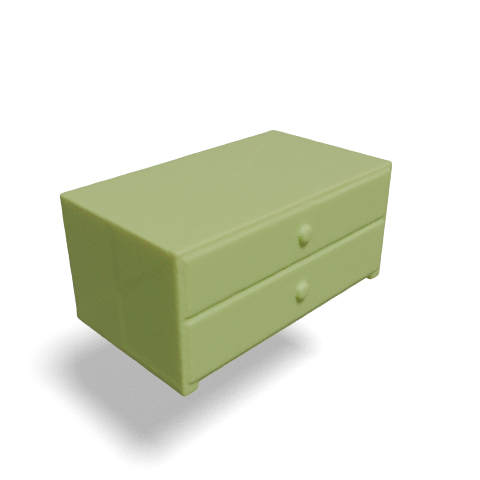} 
&\includegraphics[width=0.2\textwidth]{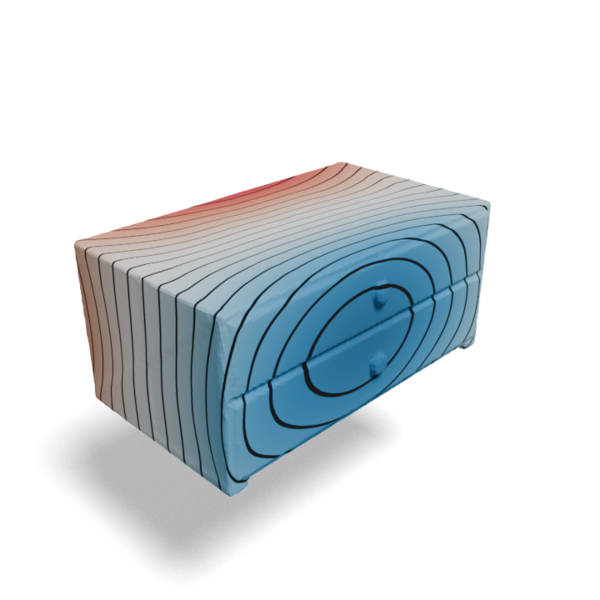} &\includegraphics[width=0.2\textwidth]{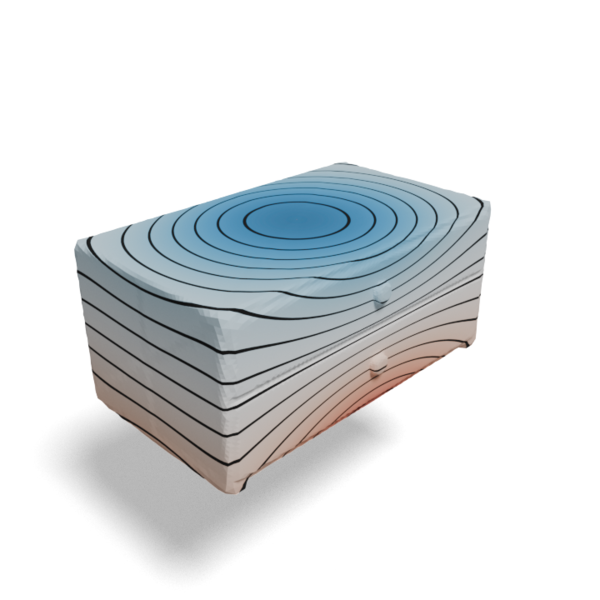}
&\includegraphics[width=0.2\textwidth]{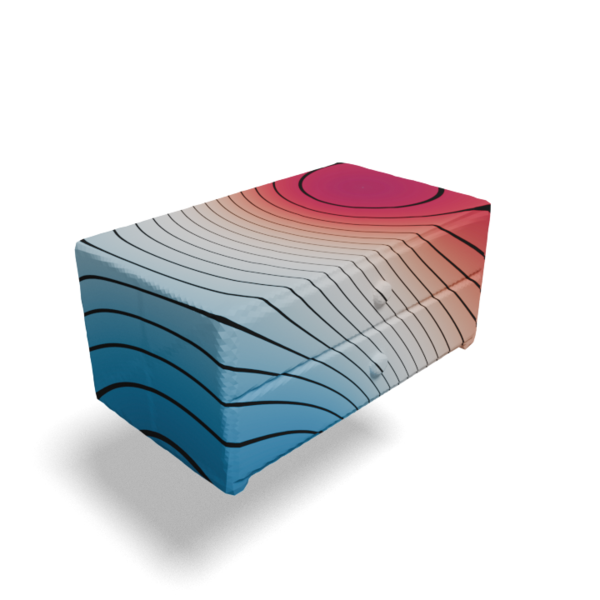} &\includegraphics[width=0.2\textwidth]{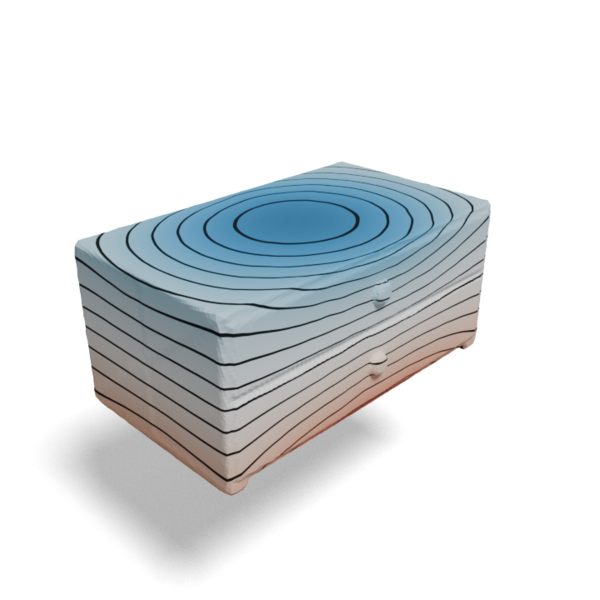} \\ 
\includegraphics[width=0.2\textwidth]{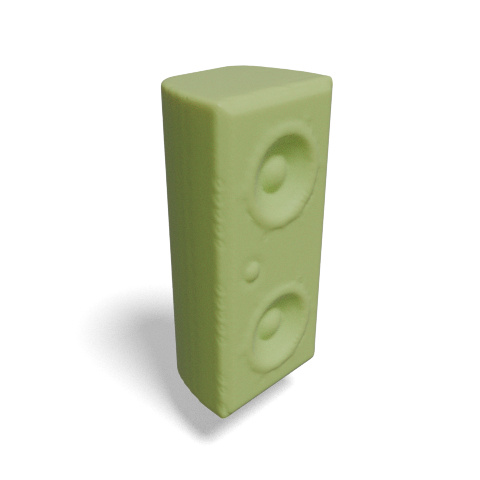} &\includegraphics[width=0.2\textwidth]{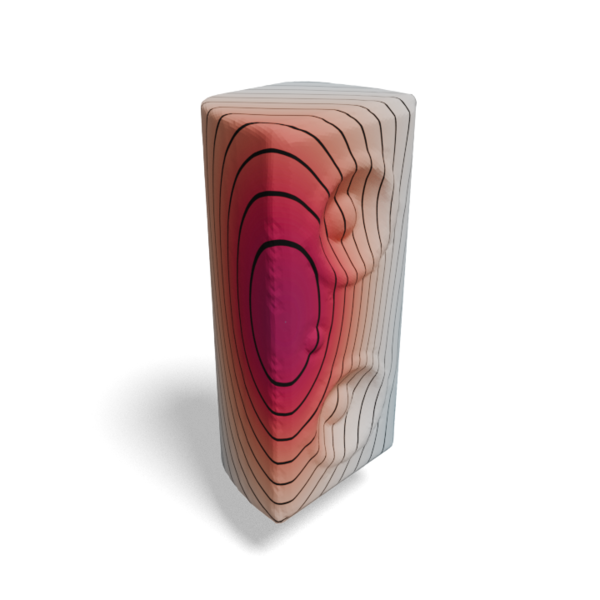}  &\includegraphics[width=0.2\textwidth]{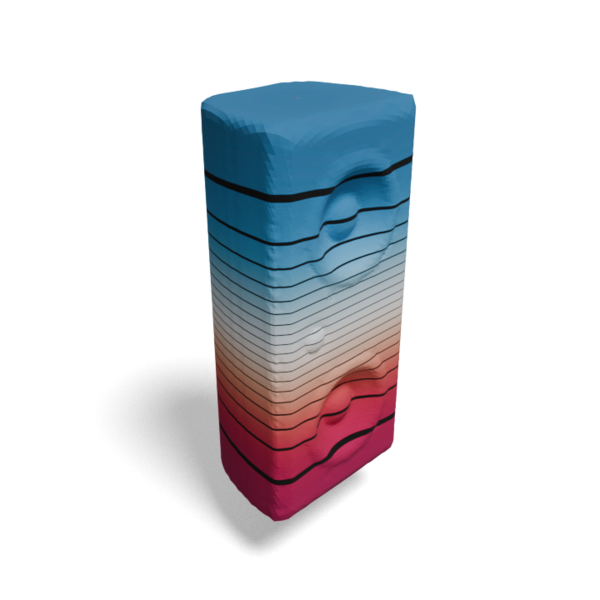}
&\includegraphics[width=0.2\textwidth]{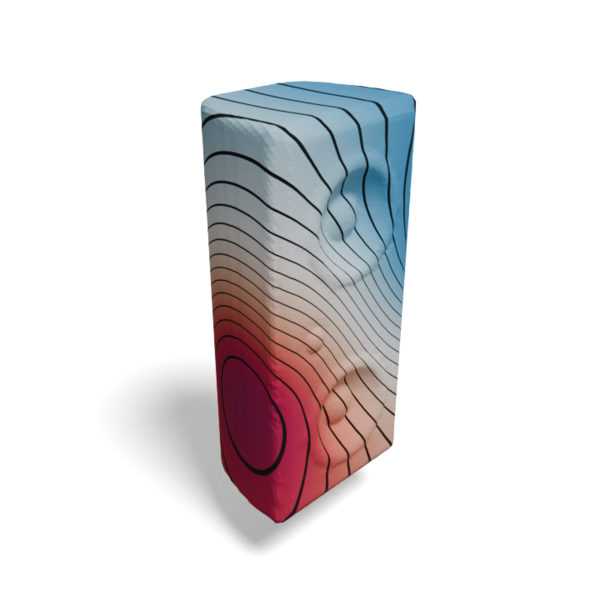} &\includegraphics[width=0.2\textwidth]{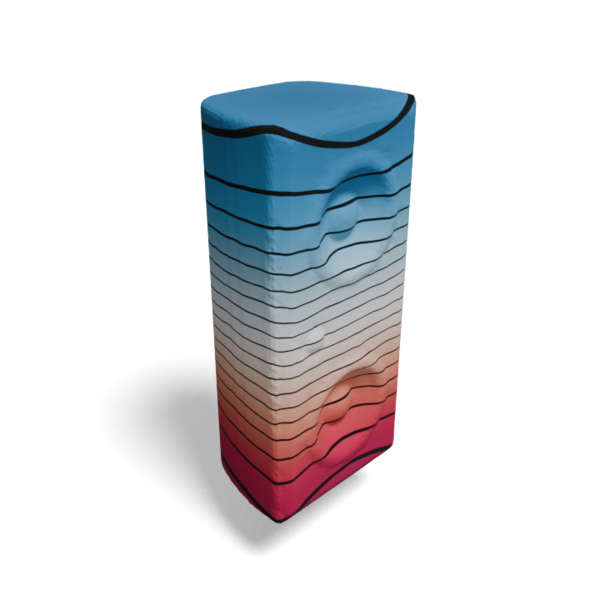} \\ \includegraphics[width=0.2\textwidth]{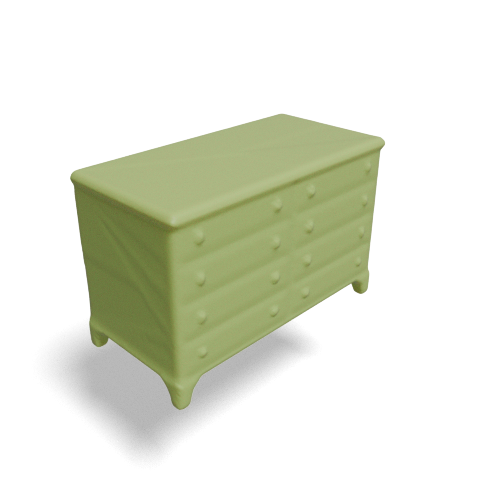} &\includegraphics[width=0.2\textwidth]{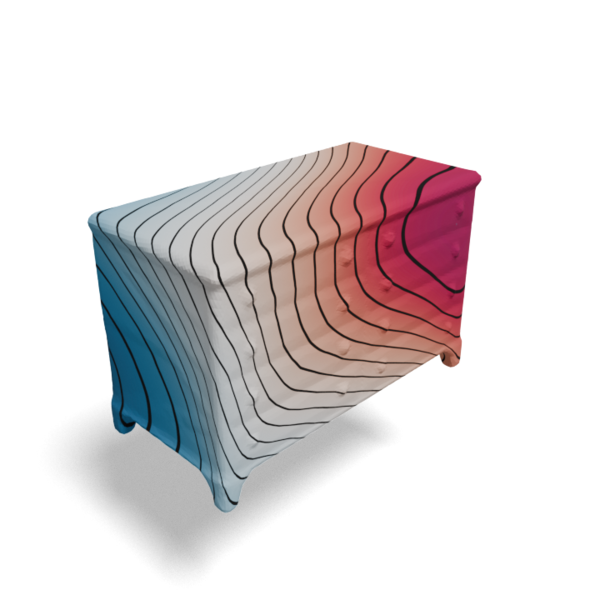} &\includegraphics[width=0.2\textwidth]{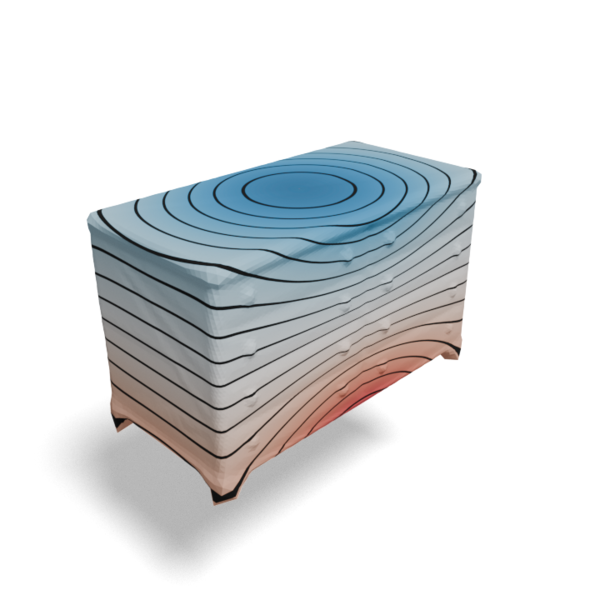}
&\includegraphics[width=0.2\textwidth]{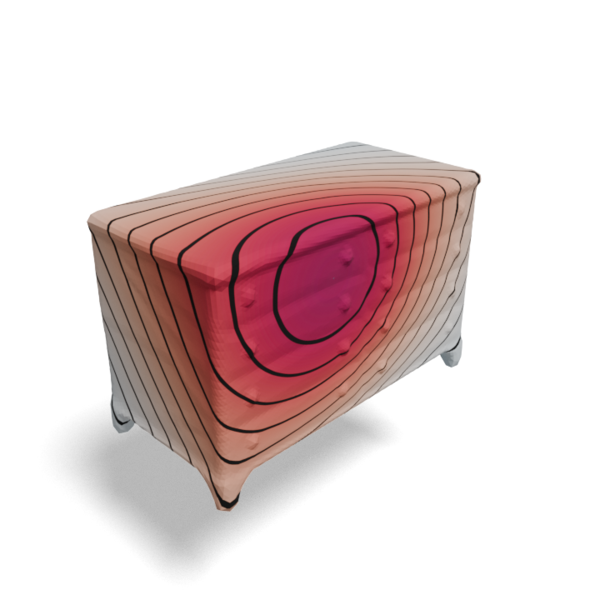} &\includegraphics[width=0.2\textwidth]{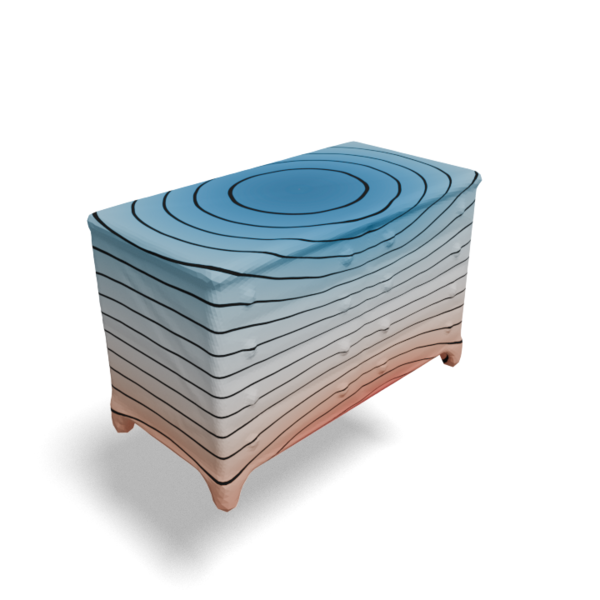} \\ \includegraphics[width=0.2\textwidth]{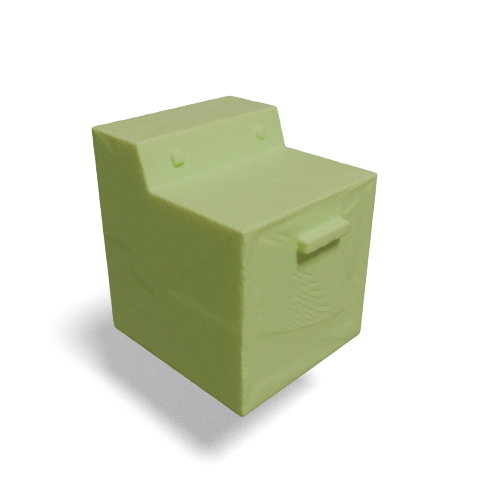} &\includegraphics[width=0.2\textwidth]{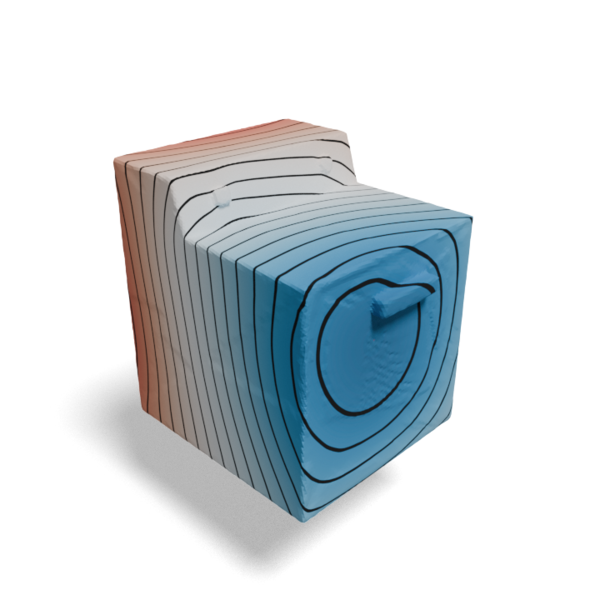}&\includegraphics[width=0.2\textwidth]{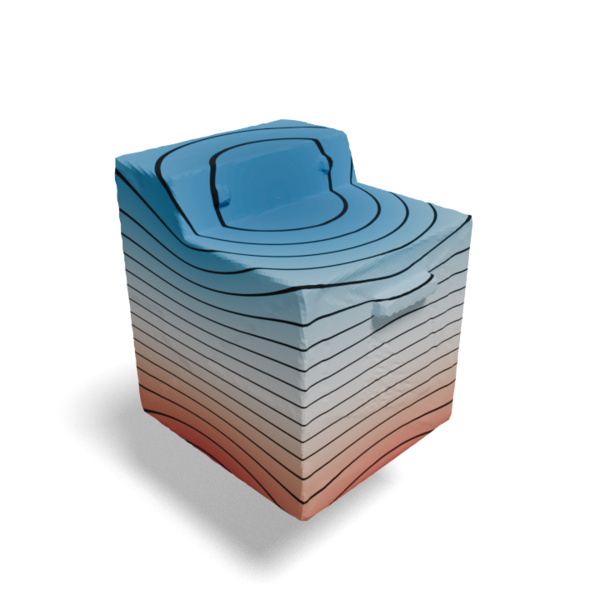}
&\includegraphics[width=0.2\textwidth]{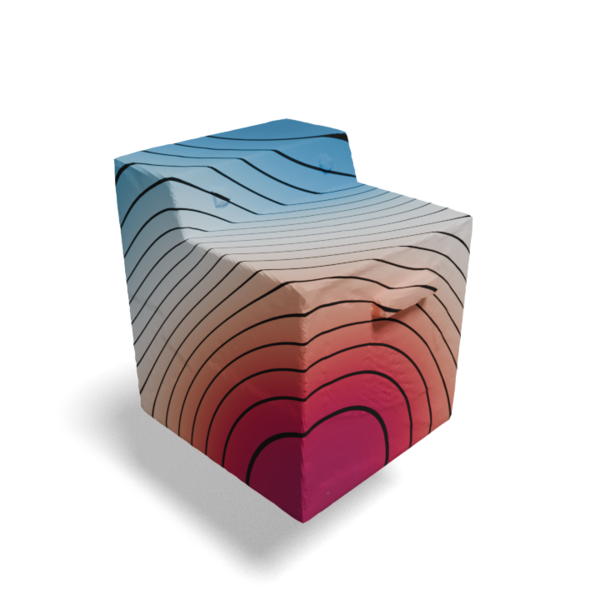} &\includegraphics[width=0.2\textwidth]{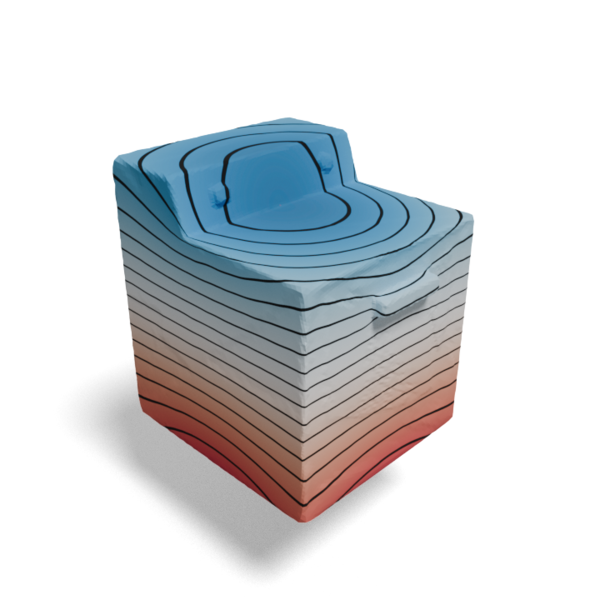}
\end{tabular}
\caption{More qualitative comparisons between GenSP and CMCF~\cite{DBLP:journals/cgf/KazhdanSB12}, SMAT ~\cite{DBLP:journals/cgf/SchmidtPK23},and AHSP~\cite{Hu-2017-AHSP}. We show images of the same sphere texture under different spherical parameterizations.}
\label{Fig:Qual:Eval2}
\end{figure*}

In Fig. \ref{Figure:Teaser}, Fig. \ref{Fig:Qual:Eval}, and Fig. \ref{Fig:Qual:Eval2}, the geometries of shapes are reconstructed from their spherical parameterizations obtained from different methods, in addition to mapping the texture defined on the sphere.

For the baseline methods which give a spherical parameterization of a shape that have the same connectivity as it, we first find the barycentric coordinates of all $40,962$ vertices of the icosphere ($\text{level}=6$) on the triangles of the given shape's spherical parameterization.
Then using these coordinates, we interpolate the vertex coordinates of the original shape while maintaining the same connectivity as that of the icosphere.

For GenSP, we first encode the ground truth shape using the shape encoder. We then use the encoded latent to deform the vertices of an icosphere with $40,962$ vertices to obtain a reconstruction of a target shape. 
While reconstructions yielded by GenSP's mesh decoder are very accurate, the generated points do not lie perfectly on the original shape. 
As the ground truth shape is available during the process of creating spherical parameterizations, we can simply project vertices generated by GenSP to the ground truth mesh.
Finally this yields a bijective map between the ico-sphere and the original shape.

We choose this reconstruction based approach for visualizing various methods of spherical parameterizations for two reasons.
Firstly, this remains consistent with the inverse spherical parameterization formulation proposed by GenSP, and secondly, it also helps to visualize the isometry and conformality  of the parameterizations through the accuracy of the reconstructions.
Due to this manner of reconstruction, we see severely under-reconstructed sections of shapes when using CMCF due to its extremely poor isometric factors.

\begin{figure*}
\centering
\setlength{\tabcolsep}{1.5pt}
\begin{tabular}{ccccccc}
GenSP &
\includegraphics[width=0.125\textwidth]{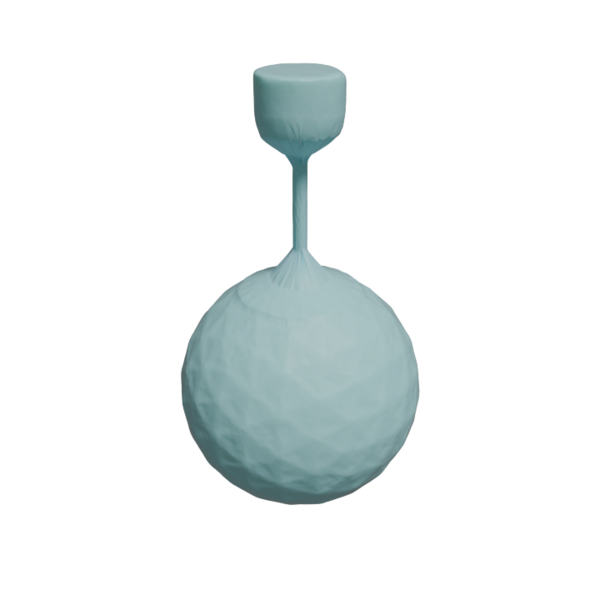} & 
\includegraphics[width=0.125\textwidth]{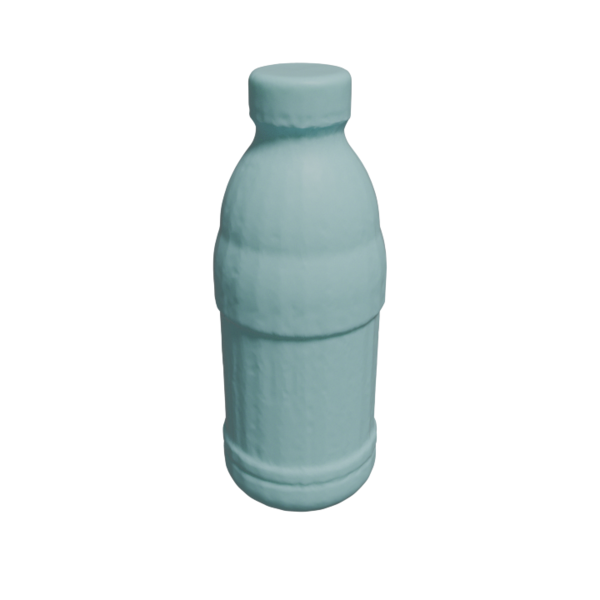} & 
\includegraphics[width=0.125\textwidth]{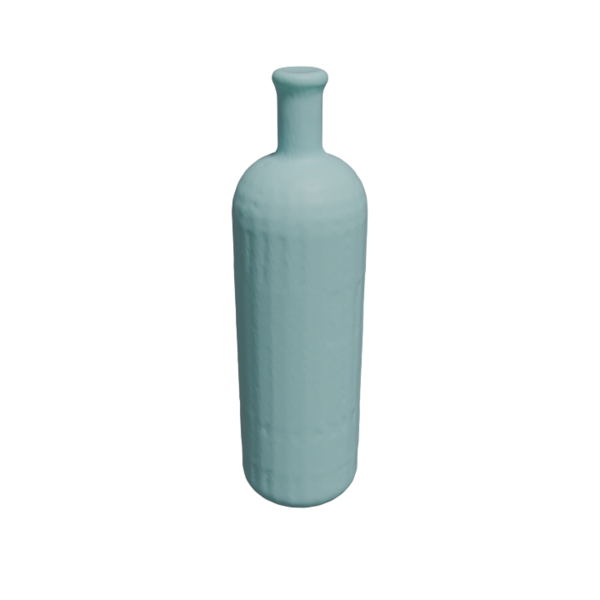} &
\includegraphics[width=0.125\textwidth]{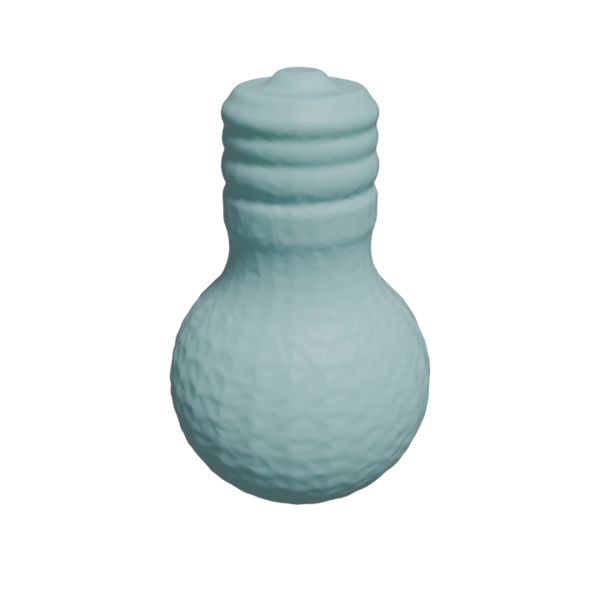} &
\includegraphics[width=0.125\textwidth]{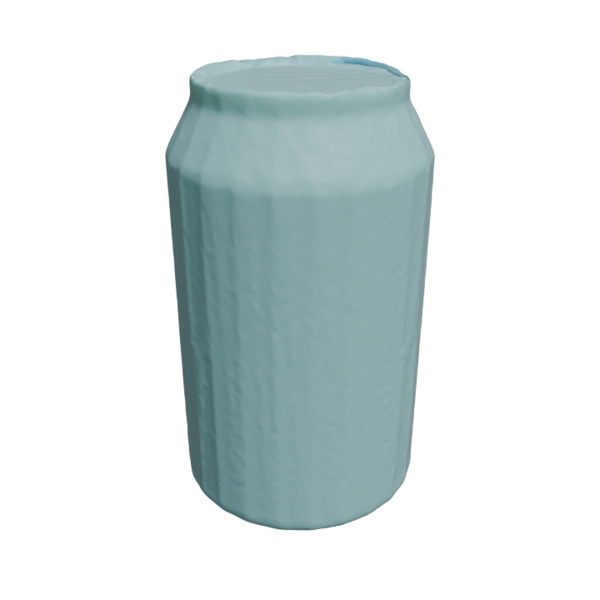} &
\includegraphics[width=0.125\textwidth]{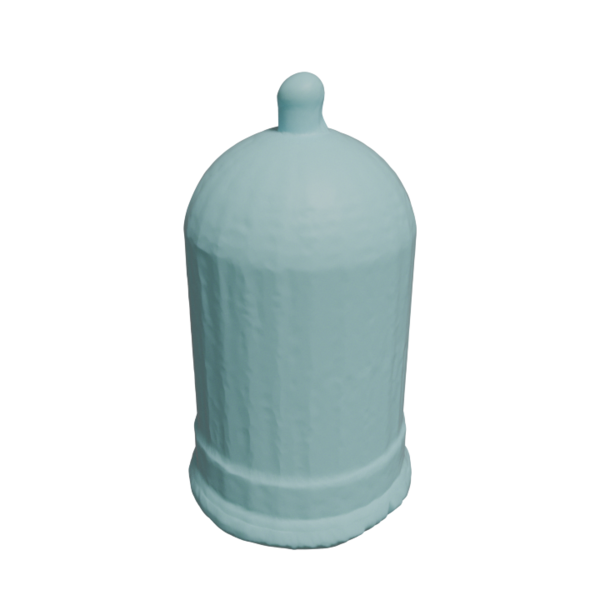} \\
No-DA & 
\includegraphics[width=0.125\textwidth]{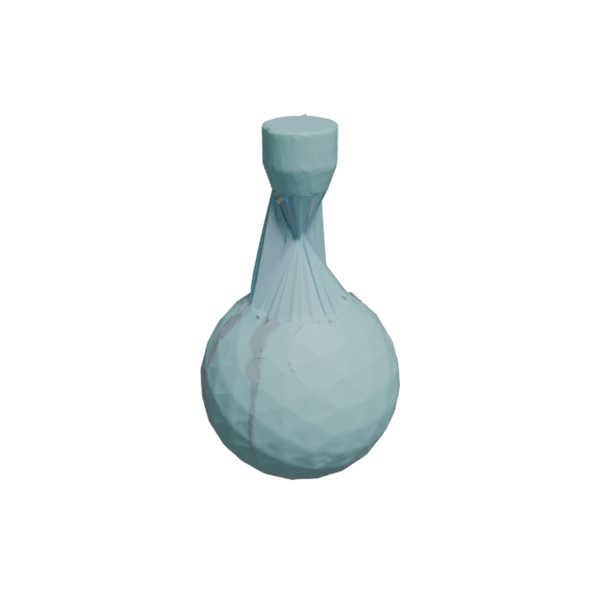} & 
\includegraphics[width=0.125\textwidth]{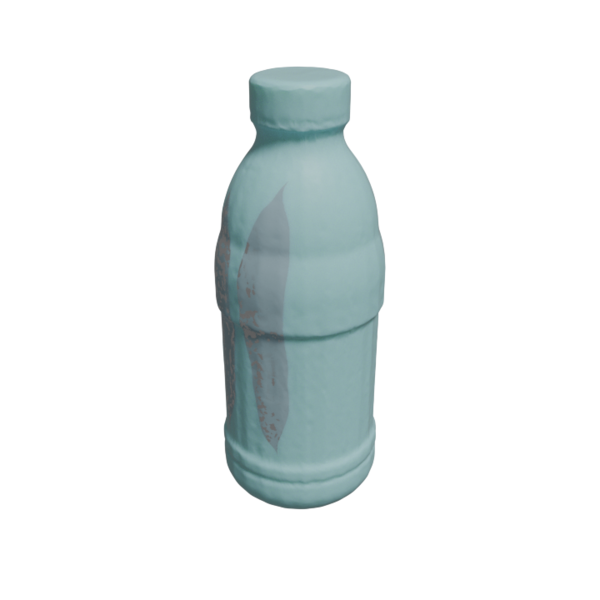} & 
\includegraphics[width=0.125\textwidth]{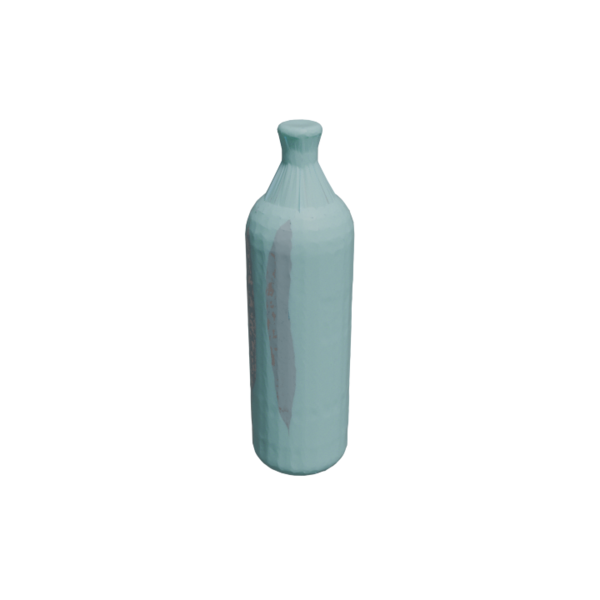} &
\includegraphics[width=0.125\textwidth]{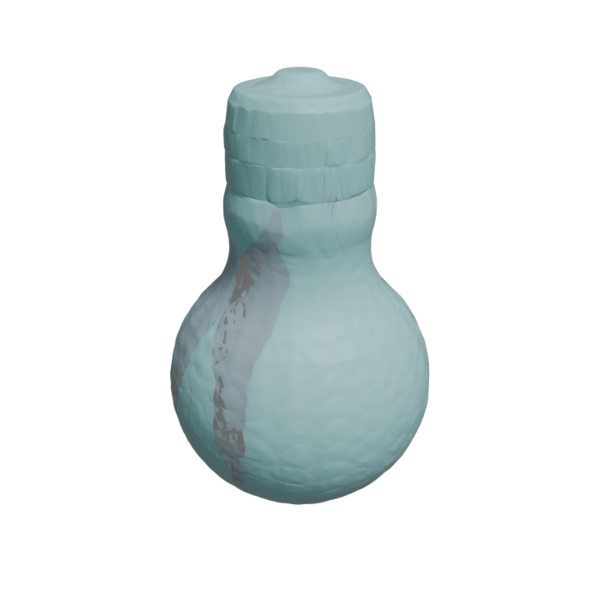} &
\includegraphics[width=0.125\textwidth]{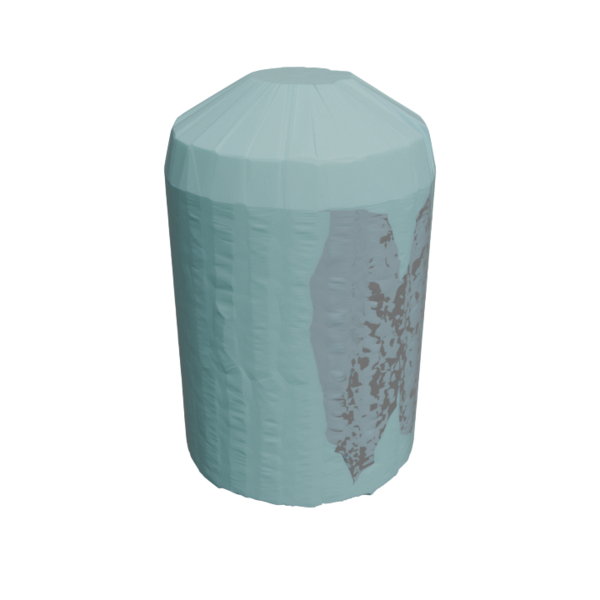} &
\includegraphics[width=0.125\textwidth]{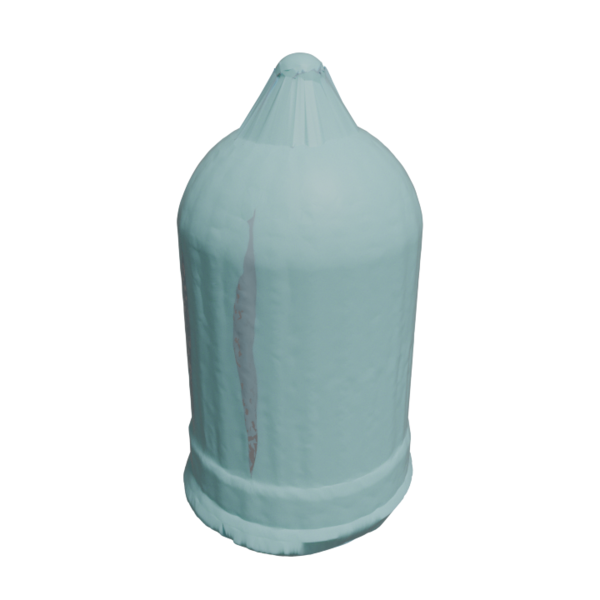}
\\
No-GP & 
\includegraphics[width=0.125\textwidth]{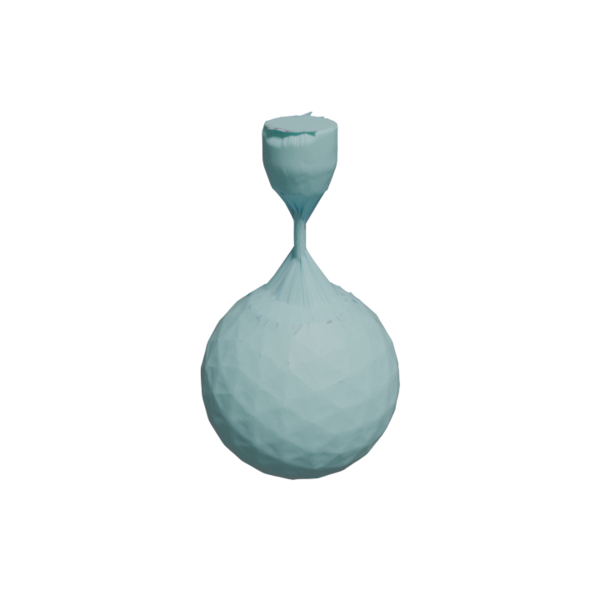} & 
\includegraphics[width=0.125\textwidth]{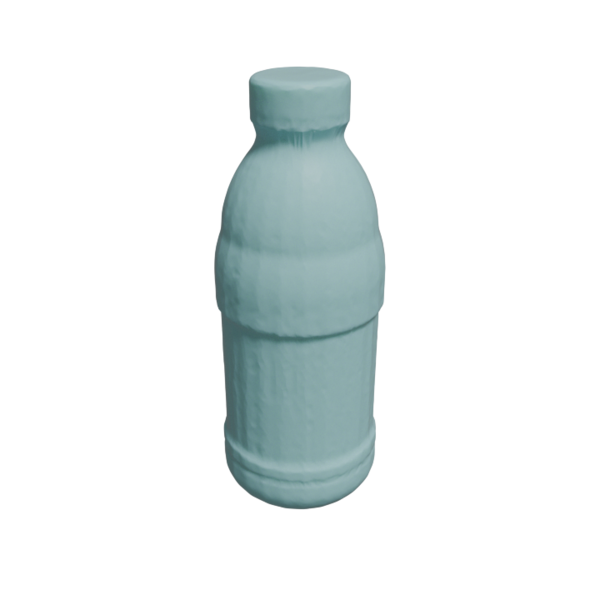} & 
\includegraphics[width=0.125\textwidth]{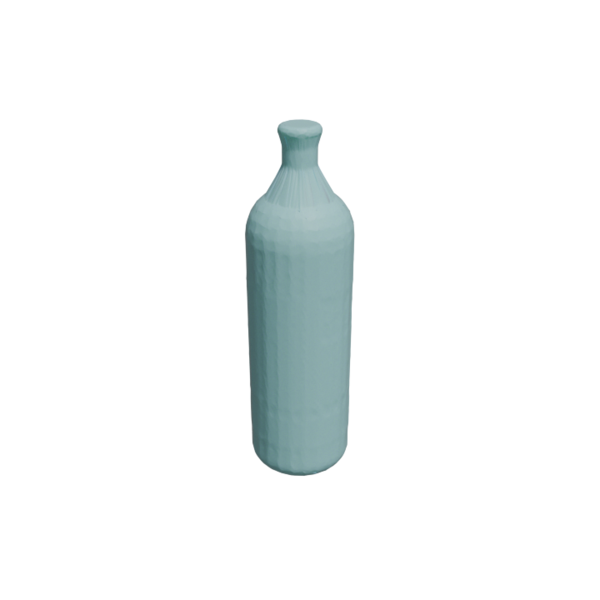} &
\includegraphics[width=0.125\textwidth]{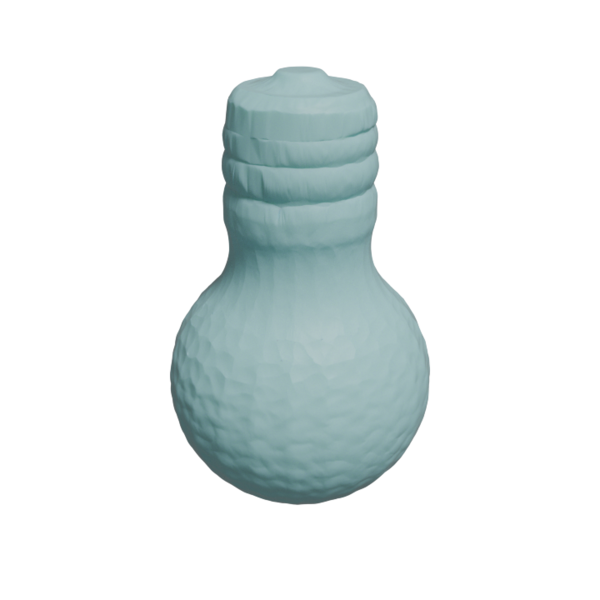} &
\includegraphics[width=0.125\textwidth]{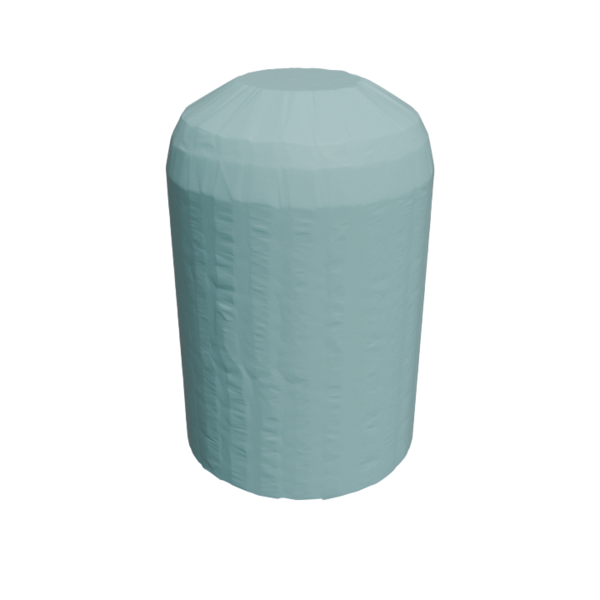} &
\includegraphics[width=0.125\textwidth]{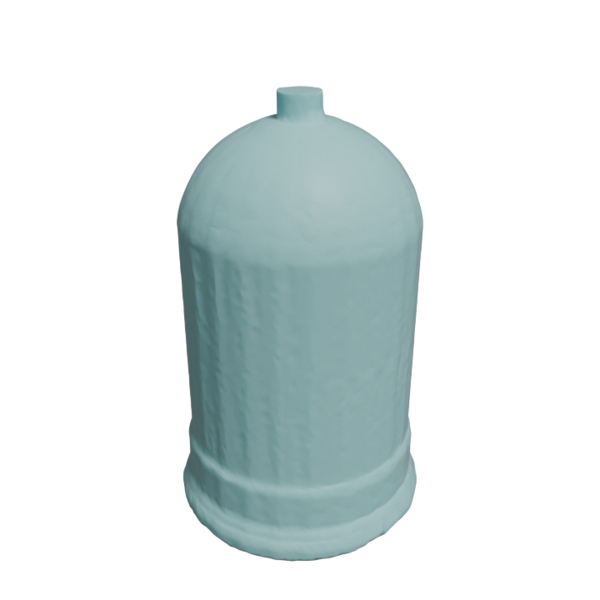} 
\\
No-GR & 
\includegraphics[width=0.125\textwidth]{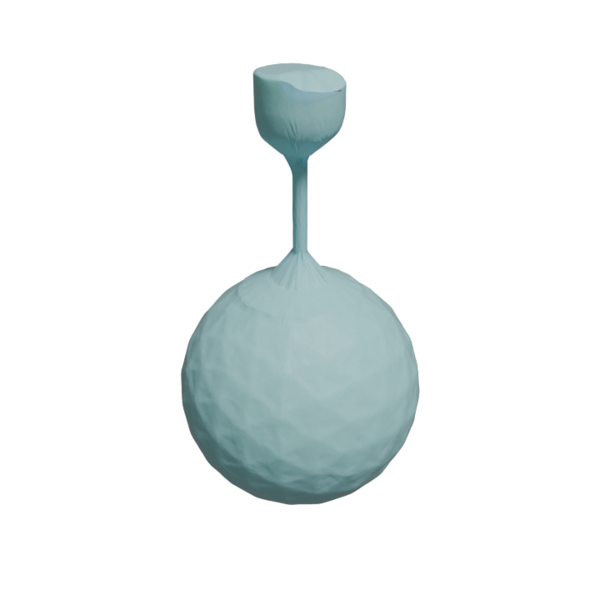} & 
\includegraphics[width=0.125\textwidth]{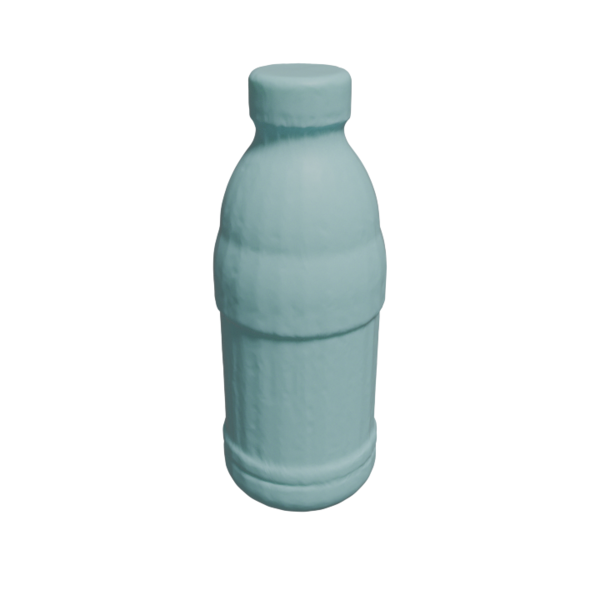} & 
\includegraphics[width=0.125\textwidth]{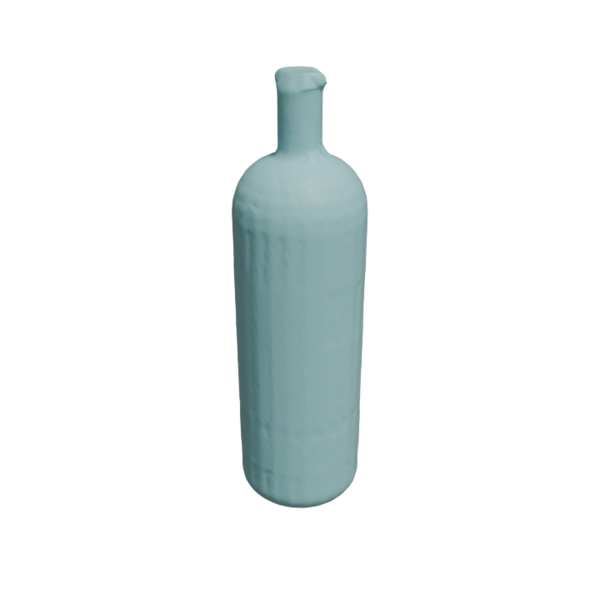} &
\includegraphics[width=0.125\textwidth]{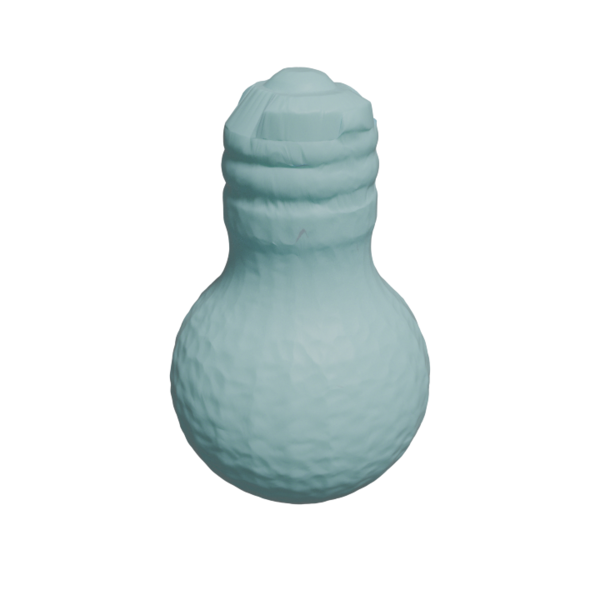} &
\includegraphics[width=0.125\textwidth]{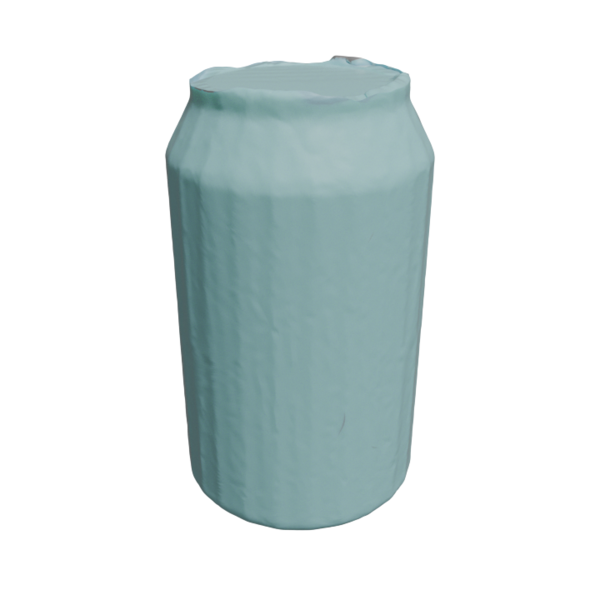} &
\includegraphics[width=0.125\textwidth]{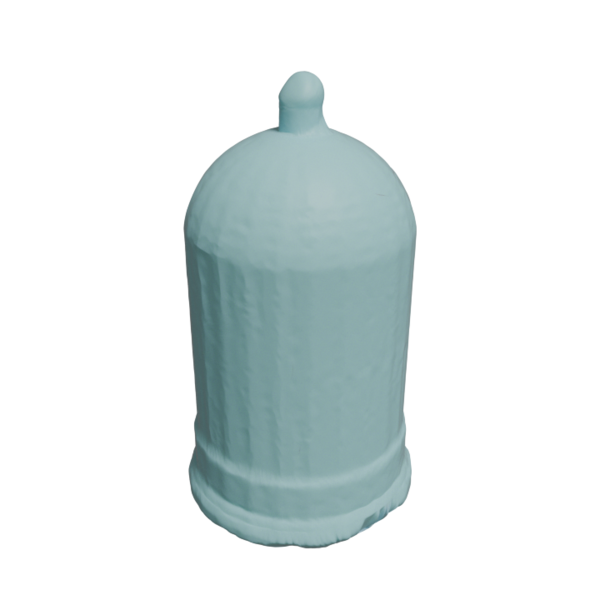} 
\end{tabular}
\caption{Qualitative ablation study results. }
\label{Fig:Ablation:Qual}
\end{figure*}

In Fig. \ref{Fig:Ablation:Qual}, we show qualitative results for removing different components of our method, as discussed in Sec. \ref{Sec:Ablation}. In the No-DA case, we observe bad geometry and major self-intersections. No-GP is perhaps the best, but struggles with thin structures and major concavities. No-GR especially struggles from bad triangulations, evident near the ridges.

\begin{figure*}
\setlength{\tabcolsep}{2pt}
\begin{tabular}{c|ccccc}
Source     & GenSP & MR & BIM & EFM \\
\hline
\includegraphics[trim={0.1cm 0.1cm 0.1cm 0.1cm}, clip, width=0.188\textwidth]{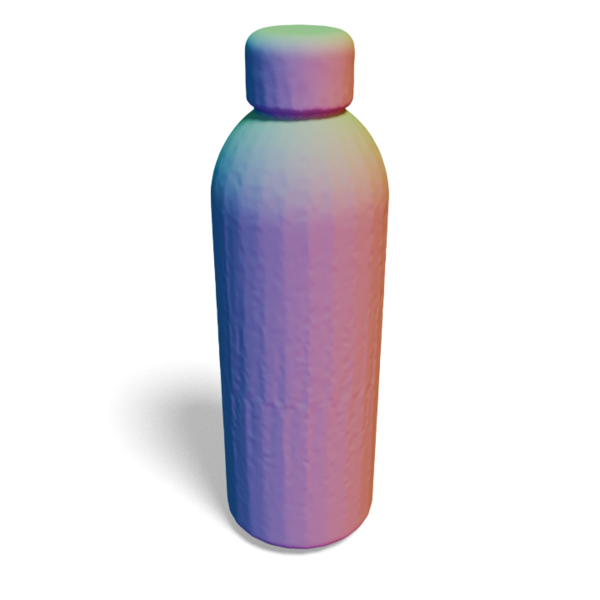}     &  \includegraphics[width=0.188\textwidth]{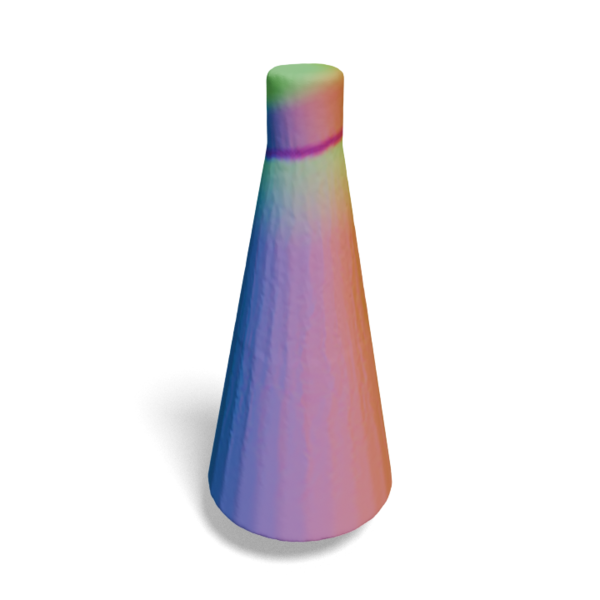}&
\includegraphics[width=0.188\textwidth]{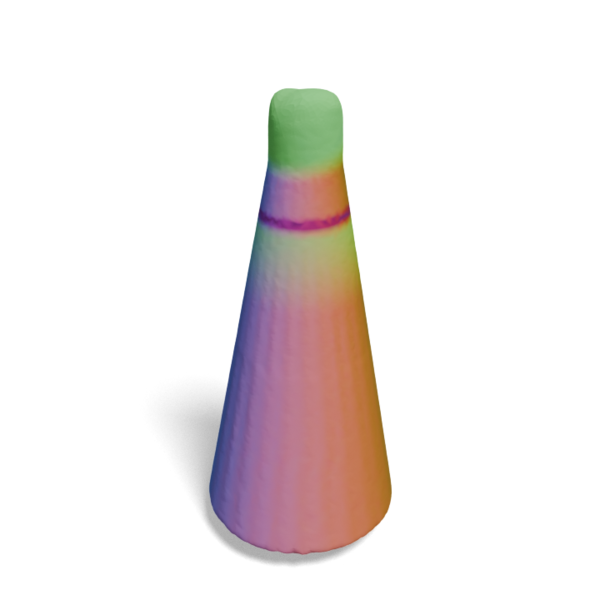}&
\includegraphics[width=0.188\textwidth]{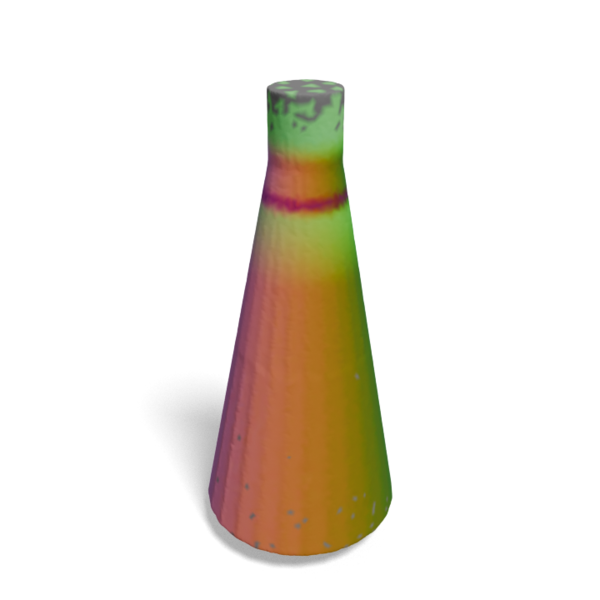}&
\includegraphics[width=0.188\textwidth]{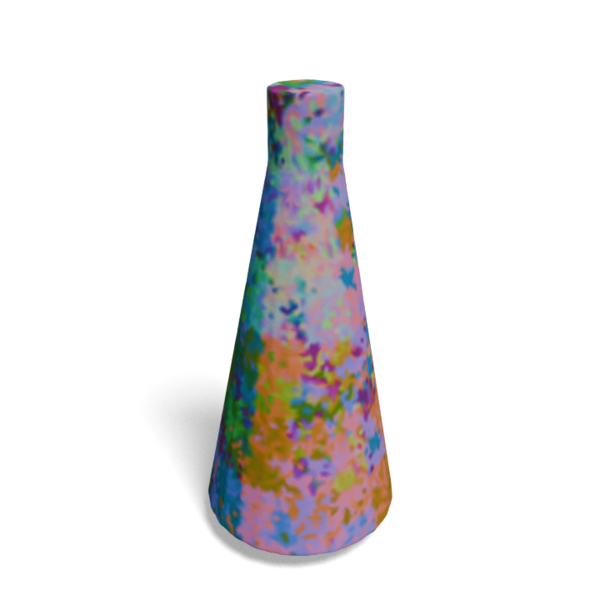}
\end{tabular}
\caption{Application in inter-shape correspondences. Baselines are Mobius registration (MR), blended intrinsic maps (BIM), and elastic functional maps (EFM). Note that EFM needs an initial set of correspondences for initialization, which is why it fails in a zero-shot setting.}
\label{Fig:Inter:Shape:Matching}
\end{figure*}

\begin{figure*}
\includegraphics[width=\textwidth]{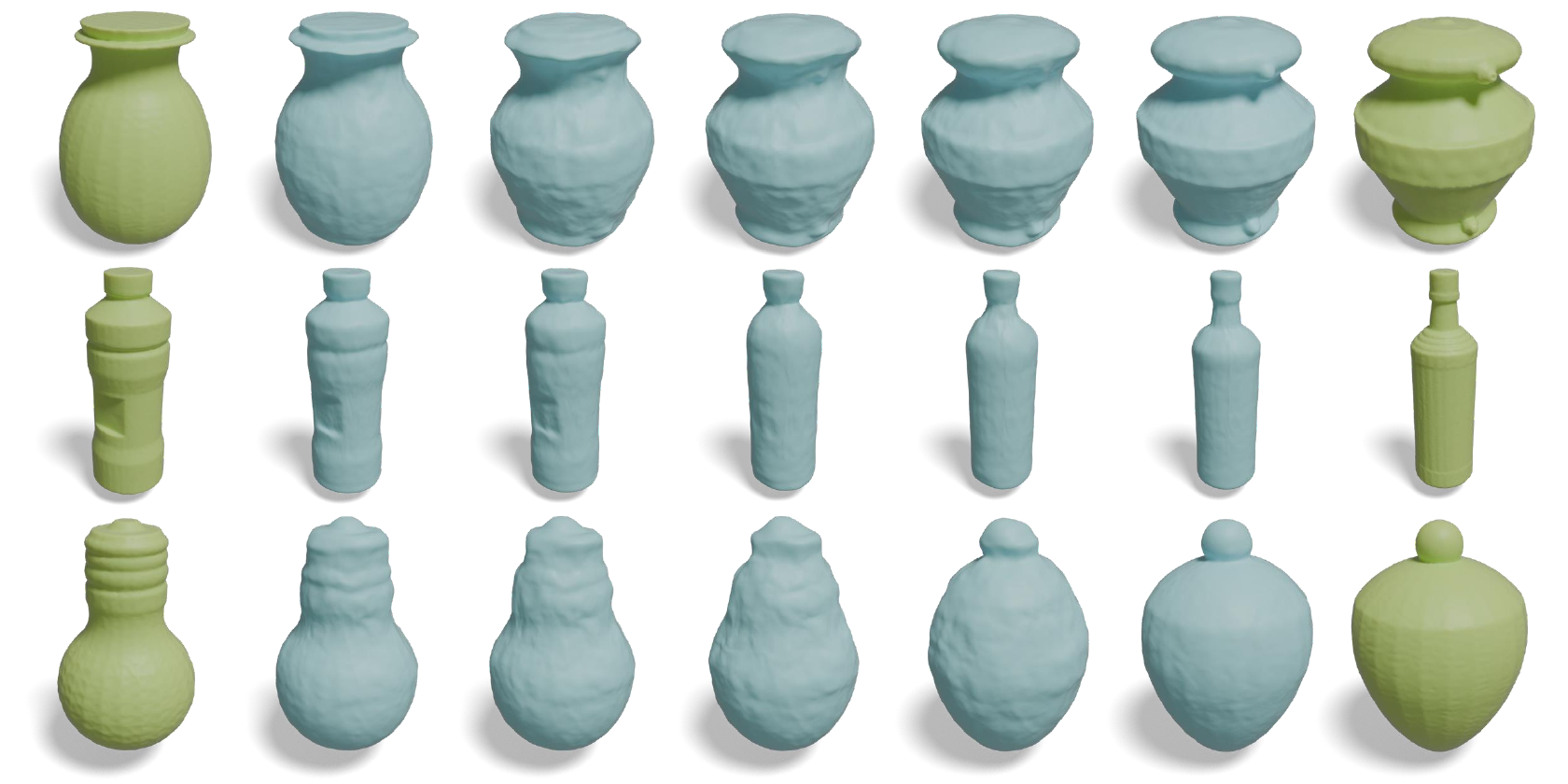}
\caption{Application in shape morphing. Applying GenSP's decoder on interpolated latent vectors produces smoothly morphed shapes. }
\label{Fig:morph}
\end{figure*}

We also demonstrate the applicability of GenSP for shape morphing and shape matching in Fig. \ref{Fig:morph} and Fig. \ref{Fig:Inter:Shape:Matching} respectively. In Fig. \ref{Fig:Inter:Shape:Matching} we compare against more classical methods like M\"obius Registration \cite{baden2018mobius}, BIM \cite{kim2011blended}, and EFM \cite{hartwig2023elastic}. 

\section{Limitations}
\begin{figure*}
    \centering
    \includegraphics[width=0.7\textwidth]{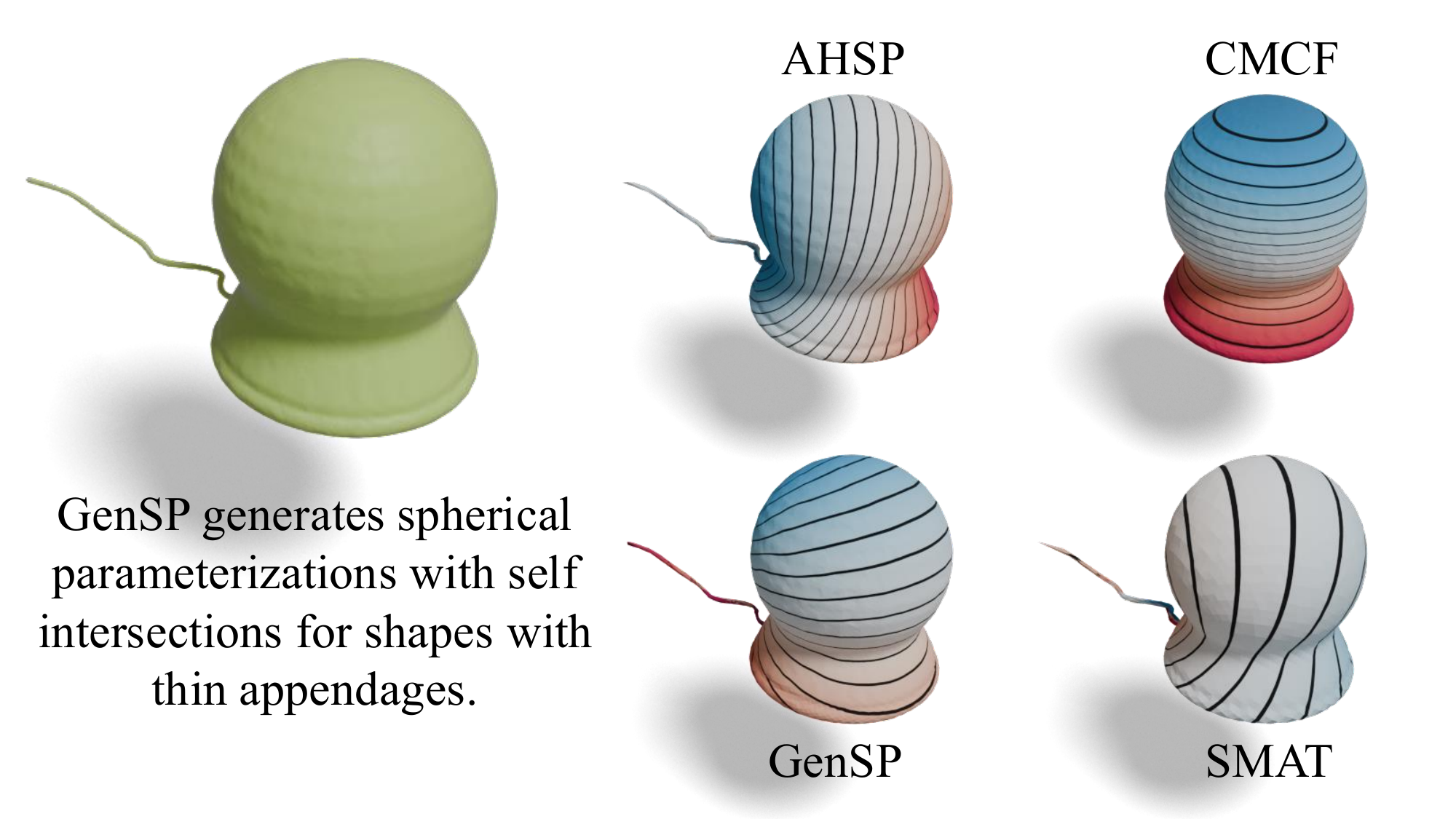}
    \caption{An example of GenSP's failure mode.}
    \label{Fig:Fail}
\end{figure*}

All methods of spherical parameterization, including ours and the baselines shown in this paper, fail for shapes with thin structures like the one shown in Fig. \ref{Fig:Fail}.
Developing algorithms that address this limitation is an interesting direction for future research.

\end{document}